%% file: 0.main.tex
\documentclass[sigconf,nonacm]{acmart}
\usepackage{multirow}
\usepackage{subfigure}
\usepackage{bm}
\usepackage{multirow}
\usepackage{subfigure}
\usepackage[normalem]{ulem}
\useunder{\uline}{\ul}{}
\usepackage{subfigure}
\usepackage{algorithm}
\usepackage[noend]{algpseudocode}
\usepackage{cite}
\usepackage{enumitem}
\AtBeginDocument{%
  \providecommand\BibTeX{{%
    \normalfont B\kern-0.5em{\scshape i\kern-0.25em b}\kern-0.8em\TeX}}}

\setcopyright{acmcopyright}
\copyrightyear{2018}
\acmYear{2018}
\acmDOI{XXXXXXX.XXXXXXX}

\acmConference[Conference acronym 'XX]{Make sure to enter the correct
  conference title from your rights confirmation emai}{June 03--05,
  2018}{Woodstock, NY}
\acmPrice{15.00}
\acmISBN{978-1-4503-XXXX-X/18/06}

\begin{document}

\title{Towards Generative Modeling of Urban Flow through Knowledge-enhanced Denoising Diffusion}

\author{Zhilun Zhou}
\affiliation{%
  \institution{BNRist, Department of Electronic Engineering,  Tsinghua University}
  \city{Beijing}
  \country{China}
}
\email{zzl22@mails.tsinghua.edu.cn}

\author{Jingtao Ding}
\orcid{0000-0001-7985-6263}
\affiliation{%
  \institution{BNRist, Department of Electronic Engineering, Tsinghua University}
  \city{Beijing}
  \country{China}
}
\email{dingjt15@tsinghua.org.cn}

\author{Yu Liu}
\affiliation{%
  \institution{BNRist, Department of Electronic Engineering, Tsinghua University}
  \city{Beijing}
  \country{China}
}
\email{liuyu2419@126.com}

\author{Depeng Jin}
\orcid{0000-0003-0419-5514}
\affiliation{%
  \institution{BNRist, Department of Electronic Engineering, Tsinghua University}
  \city{Beijing}
  \country{China}
}
\email{jindp@tsinghua.edu.cn}

\author{Yong Li}
\orcid{0000-0001-5617-1659}
\affiliation{%
  \institution{BNRist, Department of Electronic Engineering, Tsinghua University}
  \city{Beijing}
  \country{China}
}
\email{liyong07@tsinghua.edu.cn}

\begin{abstract}
Although generative AI has been successful in many areas, its ability to model geospatial data is still underexplored. Urban flow, a typical kind of geospatial data, is critical for a wide range of applications from public safety and traffic management to urban planning. Existing studies mostly focus on predictive modeling of urban flow that predicts the future flow based on historical flow data, which may be unavailable in data-sparse areas or newly planned regions. Some other studies aim to predict OD flow among regions but they fail to model dynamic changes of urban flow over time. In this work, we study a new problem of urban flow generation that generates dynamic urban flow for regions without historical flow data. 
To capture the effect of multiple factors on urban flow, such as region features and urban environment, we employ diffusion model to generate urban flow for regions under different conditions. We first construct an urban knowledge graph (UKG) to model the urban environment and relationships between regions, based on which we design a knowledge-enhanced spatio-temporal diffusion model (KSTDiff) to generate urban flow for each region. Specifically, to accurately generate urban flow for regions with different flow volumes, we design a novel diffusion process guided by a volume estimator, which is learnable and customized for each region. Moreover, we propose a knowledge-enhanced denoising network to capture the spatio-temporal dependencies of urban flow as well as the impact of urban environment in the denoising process.
Extensive experiments on four real-world datasets validate the superiority of our model over state-of-the-art baselines in urban flow generation. Further in-depth studies demonstrate the utility of generated urban flow data and the ability of our model for long-term flow generation and urban flow prediction.
Our code is released at: https://github.com/tsinghua-fib-lab/KSTDiff-Urban-flow-generation.
\end{abstract}


\maketitle

\input{1.introduction}

\input{2.related_work}

\input{3.preliminaries}

\input{4.methods}

\input{5.experiments}

\input{6.conclusion}

\bibliographystyle{ACM-Reference-Format}
\bibliography{reference}

\newpage
\appendix

\input{7.Appendix.tex}

\end{document}

%% file: 1.introduction.tex
\section{Introduction}

Generative AI, especially large pre-trained models, have shown great success in many areas such as natural language processing and computer vision, with recently popular GPT-4~\citep{openai2023gpt4} and Stable Diffusion~\citep{rombach2022high} as examples. However, their ability to model geospatial data is still underexplored~\citep{mai2022towards}.

\begin{figure}[htbp!]
\vspace{-5px}
\centering
\includegraphics[width=.9\linewidth]{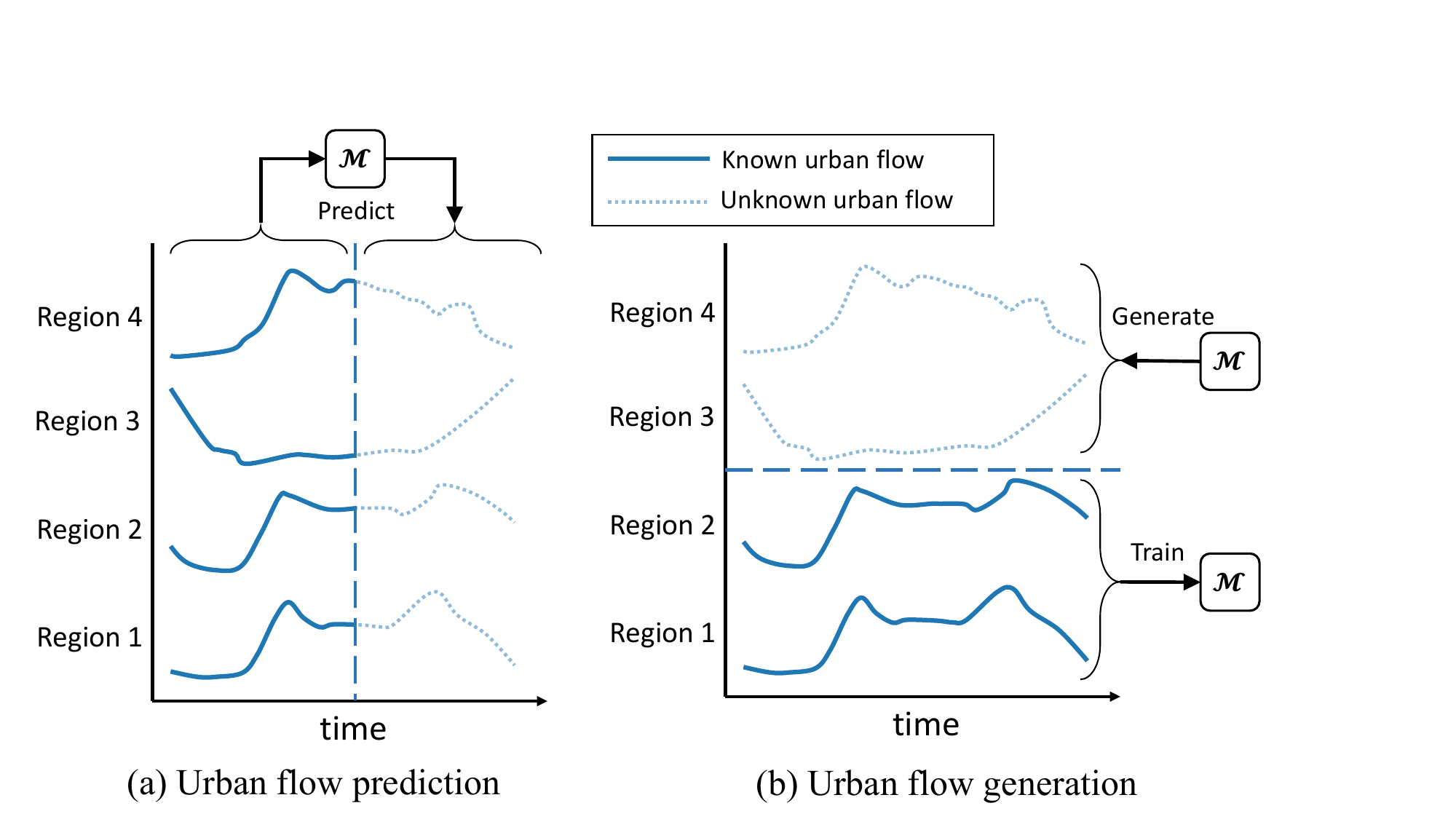}
\vspace{-5px}
\caption{Comparison of urban flow prediction and urban flow generation.}
\label{fig:pred_vs_gen}
\vspace{-10px}
\end{figure}

Urban flow is a typical kind of geospatial data that depicts the dynamic human mobility patterns in urban regions~\citep{wang2022generative}, playing an important role in public safety, traffic management, and urban planning. For example, the inflow and outflow refer to the number of people entering or leaving an urban region during a given time interval. With such data, the government can implement preventive measures in advance to avoid stampede or traffic congestion in regions experiencing a large inflow. Moreover, ride-sharing platforms can dispatch more taxis to regions with large outflows to meet the demand. 
In recent years, many studies have been conducted on \textit{predictive modeling} of urban flow. As shown in Figure~\ref{fig:pred_vs_gen}(a), they mostly focus on urban flow prediction, which trains a prediction model based on historical flow data to predict the future urban flow of regions~\citep{liu2022msdr,lin2019deepstn+,zhang2017deep}. However, these studies rely heavily on historical flow, and thus cannot be adapted to data-sparse areas such as newly planned regions or suburban regions.
In addition, some studies have made efforts to predict static origin-destination flows from one region to another given their characteristics~\citep{zipf1946p,simini2021deep,liu2020learning}. Nevertheless, they fail to capture temporal patterns since they cannot generate dynamic urban flow data, where the flow may vary significantly at different times of the day.

In this work, we take a step towards \textit{generative modeling} of urban flow by studying a new problem of urban flow generation, which aims to train a generation model based on existing regions in order to estimate the urban flow of new regions without historical flow data, as shown in Figure~\ref{fig:pred_vs_gen}(b). Urban flow generation can help estimate the urban flow for data-sparse regions or evaluate the flow patterns of newly planned regions in advance of region construction, which is essential for urban planning~\citep{zhang2020curb}.

The urban flow of a region is affected by multiple factors including regional characteristics, urban environment, and urban flow in nearby regions. Therefore, urban flow generation can be formulated as a conditional generation problem, which aims to generate flow for different regions under different conditions. Recently, diffusion models have shown outstanding performance in a wide range of tasks like image synthesis~\citep{ho2020denoising}, audio synthesis~\citep{kong2020diffwave}, and time series modeling~\citep{tashiro2021csdi,rasul2021autoregressive}. The key idea of diffusion models is to first gradually add noise to data until it becomes random noise, and then train a model to learn the reverse process, i.e., convert random noise into expected data distribution step by step. Moreover, additional information can be injected into each step of the reverse process, making it easy to generate data under given conditions using diffusion models~\citep{yuan2023spatio}.
As a result, we propose to leverage conditional diffusion models to solve the urban flow generation problem so as to better control the generation process by multiple factors.

\begin{figure}[htbp!]
\centering
\vspace{-5px}
    \subfigure[Washington, D.C.]{
    {\label{subfig:dc_scale}}
    \includegraphics[width=.4\linewidth]{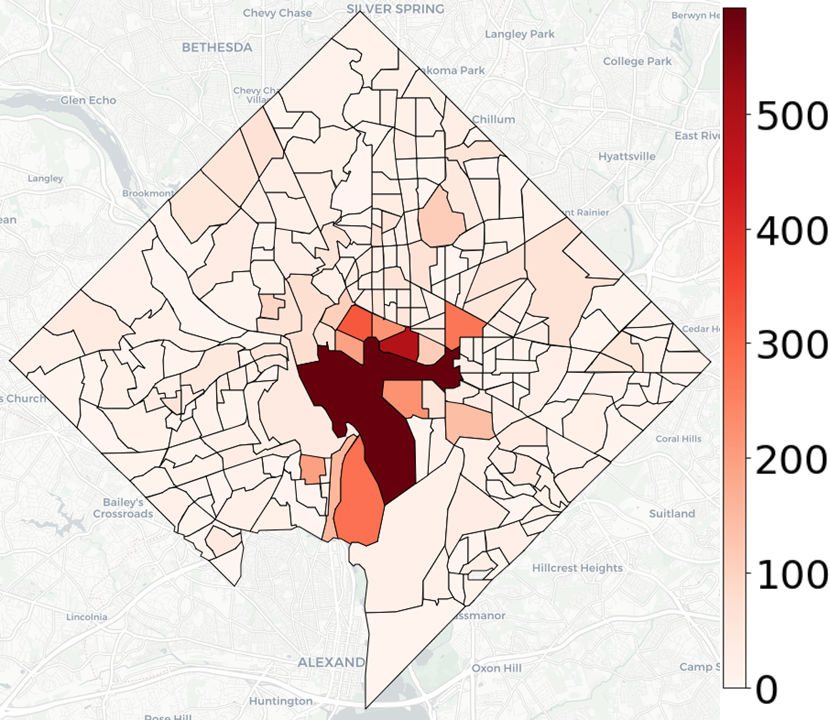}
    }
    \subfigure[Baltimore]{
    {\label{subfig:bm_scale}}
    \includegraphics[width=.4\linewidth]{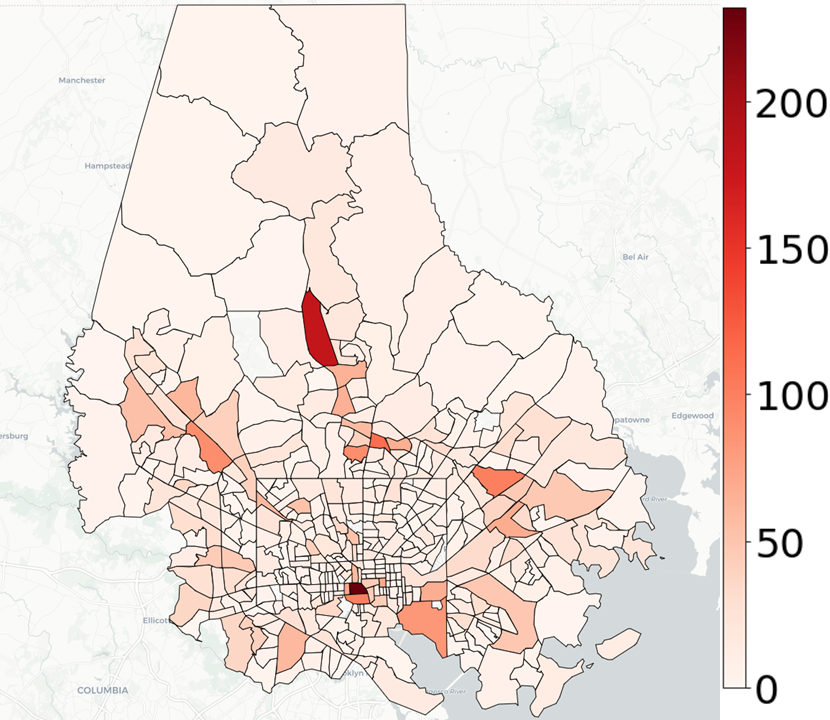}
    }
\vspace{-10px}
\caption{The average flow volume varies significantly in different regions.}
\label{fig:scale}
\vspace{-5px}
\end{figure}

However, accurately generating urban flow under diverse conditions still faces the following challenges:
(1) \textbf{Significant variance in the spatial distribution of urban flow volume}. The volume of urban flow varies significantly across regions~\citep{ji2022spatio}. For example, newly planned regions located in suburban areas typically exhibit lower volumes than established regions in downtown areas due to underdeveloped infrastructure and transportation networks.
We visualize the flow volume in different regions of Washington, D.C., and Baltimore, which is defined as the average flow in each hour of a region. As shown in Figure~\ref{fig:scale}, the flow volumes of regions in downtown areas are generally much larger than those in suburban areas. Such spatial heterogeneity in flow volume distribution makes it a challenging task to generate urban flow accurately for different regions with different volumes without historical flow data. 
(2) \textbf{Highly complex and varying spatio-temporal dependencies or urban flow across different regions}. There exist complex spatio-temporal dependencies of urban flow among regions, and such dependencies are affected by urban environment and vary across different regions. For example, in downtown business areas with dense road networks, the urban flow of a region may be largely influenced by nearby regions, resulting in strong spatial correlations. On the contrary, for residential regions in suburban areas, the flow transition among regions may be rather small, and urban flow there exhibits stronger temporal patterns, e.g., large outflow during weekday morning and large inflow at night. Most of the existing diffusion models can only model temporal correlation in time series~\citep{rasul2021autoregressive,tashiro2021csdi}, while they fail to consider the spatio-temporal dependencies in urban flow and the effect of urban environment.

To tackle the complexity of modeling urban flow and regions, we construct an urban knowledge graph (UKG) to model the urban environment as well as various relationships among regions. Based on UKG, we propose a \textbf{K}nowledge-enhanced \textbf{S}patio-\textbf{T}emporal \textbf{D}iffusion model~(KSTDiff) to generate dynamic urban flow for regions.
Specifically, in the forward diffusion process, we gradually add Gaussian noise to the urban flow data of each region. To handle the variance of flow volume among regions, we design a volume estimator to directly estimate the flow volume for each region, which is then used to guide the diffusion process by adjusting the mean of added Gaussian noise accordingly. Moreover, the volume estimator is jointly trained with the diffusion model to make the diffusion process of each region learnable and customized, and thus overcome the first challenge.
In the denoising process, we aim to train a denoising network to convert random noise to urban flow data step by step. To solve the second challenge, we design a knowledge graph enhanced spatio-temporal block (KGST block) to capture spatio-temporal dependencies of urban flow based on environmental information, where we utilize relation-aware graph convolution layers to capture multiple types of spatial dependencies based on UKG and use transformer layers to model temporal dependencies. Moreover, we leverage KG embeddings of regions with environmental information preserved to guide the spatio-temporal fusion. In this way, we manage to capture the spatio-temporal dependencies as well as the environmental factors that affect such dependencies.

Overall, our contributions can be summarized as follows:
\begin{itemize}[leftmargin=10px]
\item We study the problem of urban flow generation, which aims to generate urban flow for regions without historical flow data, by proposing a knowledge-enhanced spatio-temporal diffusion model. To our best knowledge, we are the first to introduce diffusion model for generative modeling of geospatial data.
\item We employ diffusion model to control the urban flow generation process based on multiple factors. Specifically, we design a volume estimator to guide the diffusion process with estimated flow volume for each region separately. Moreover, we construct an urban KG to model urban environment as well as complex relationships among regions, and devise an STKG block to capture spatio-temporal dependencies of urban flow.
\item We conduct extensive experiments on four real-world urban flow datasets, and our model outperforms state-of-the-art models by over 13.3\% in terms of MAE on all datasets, highlighting the superiority and robustness of our model. Several in-depth studies further demonstrate the effectiveness of our model design and its ability to generate long-term flow data. Moreover, experiments also demonstrate the utility of generated data for downstream applications and our model's capability for predictive modeling of urban flow.
\end{itemize}

%% file: 2.related_work.tex
\section{Related Work}
\label{sec:related work}

\subsection{Urban Flow Modeling}
Understanding urban flow is crucial for many areas like urban planning and transportation system. In recent years, plenty of works have studied the problem of urban flow prediction through deep learning methods~\citep{zhang2017deep,jin2023spatio,li2022lightweight}, which leverage neural networks to predict the future inflow and outflow for citywide regions based on historical flow. The core of urban flow prediction is to model the spatio-temporal dependencies of urban flow. Existing works mostly use convolutional neural networks (CNN) or graph neural networks (GNN) to capture spatial dependencies, and use recurrent neural networks (RNN) or transformers to capture temporal dependencies.
However, these works depend on historical flow and thus cannot be used for new regions without such data.

Another series of works aims to predict OD flow among regions. Deep Gravity~\citep{simini2021deep} leverages neural networks to predict OD flow among each pair of regions based on their features, distance as well as outflows. GMEL~\citep{liu2020learning} first learns an embedding for each region through graph attention network (GAT), and then use regression model to estimate the OD flow based on region embeddings. 
However, these works can only generate static urban flow, which is not enough for modeling urban flow with dynamic changes over time.

\subsection{Diffusion Model for Time Series Modeling}

Diffusion models have been widely used in time series forecasting, imputation, and generation owing to their ability to model high dimensional data distributions.
For example, TimeGrad~\citep{rasul2021autoregressive} is used for time series forecasting in an autoregressive way. It combines diffusion model with RNN and generates the value of the next time step conditioned on the hidden state of RNN.
CSDI~\citep{tashiro2021csdi} uses diffusion models for probabilistic time series imputation by generating missing values conditioned on observed values.
Diffusion models have also been used in time series generation applications like electronic health records (EHR) synthesis~\citep{alcaraz2023diffusion,he2023meddiff,yuan2023ehrdiff}.

Most of these works use the denoising network architecture proposed in Diffwave~\citep{kong2020diffwave}, which leverages bidirectional dilated convolution to capture the correlation between different time steps.
However, these works fall short of modeling spatial dependencies among different regions in our urban flow generation task. 
Moreover, a recent work named DiffSTG~\citep{wen2023diffstg} designs a novel UGnet for spatio-temporal graph forecasting, which can capture spatio-temporal dependencies of different nodes, while it depends on historical data to learn such dependencies and cannot manage to generate urban flow without historical flow data.

%% file: 3.preliminaries.tex
\section{Preliminaries}
\label{sec:preliminaries}
\subsection{Problem Statement}
In this section, we formally define our problem based on some concepts, and give a brief introduction of diffusion model. The notations used in the following sections are summarized in Table~\ref{tbl:notations}.
\begin{definition}[\textbf{Urban Region}]
Urban regions are defined as non-overlapping areas in a city partitioned by main road networks, or defined as administrative areas such as census tracts.
\end{definition}

\begin{definition}[\textbf{Urban Flow}]
In this study, we focus on two kinds of urban flow, namely the inflow and outflow, which are defined as the number of people entering or leaving the region in a given time interval. They are usually calculated using user trajectory data or taxi trip data.
In addition, the inflow of a region can also be calculated as the total number of visits to POIs in the region based on POI check-in data.
The urban flow at time $t$ can be represented as $\bm{F}_t\in \mathbb{R}^{N_l\times d_f}$, where $N_l$ is the number of target regions and $d_f$ is the dimension of urban flow, e.g., $d_f=2$ for inflow and outflow.
\end{definition}

Based on these concepts, we define the research problem as follows.
\begin{definition}[\textbf{Urban Flow Generation}]
Given a set of regions $\mathcal{S}_l=\{l_1,l_2,\ldots,l_{N_l}\}$ with attributes like population and POI information, generate urban flow for these regions in different time intervals $\bm{F}_1,\ldots,\bm{F}_T$ without historical flow data. The distribution of generated urban flows should be similar to real flows.
\end{definition}

The urban flow of each region is affected by many factors like region features, urban environment, and interactions among regions. To better utilize such factors for urban flow generation, we employ denoising diffusion probabilistic model (DDPM) as the backbone of our model. Moreover, we construct an urban knowledge graph (UKG) to provide a comprehensive description of urban environment as well as complex relationships among regions, and generate urban flow based on the UKG.
We first give a brief introduction to DDPM and the construction of UKG.

\begin{table}[h]
    \centering
    \caption{Notations.}
    \vspace{-10px}
    \resizebox{0.95\linewidth}{!}{
\begin{tabular}{c|l}
\hline
\textbf{Notations} & \textbf{Description}                      \\ \hline
$N_l$              & The number of regions                      \\
$T$              & The length of generated flow (number of time intervals)   \\
$d_f, d_{fea}$ & Dimension of urban flow and region features\\
$d_{KG}, d_h$ & Dimension of KG embedding and hidden size\\
$\bm{\epsilon}$         & Gaussian noise                            \\
$\bm{\epsilon}_\theta$         & Denoising module with parameter $\theta$     \\
$\hat{\bm{\epsilon}}$   & Noise predicted by denoising module       \\
$N$                & The number of diffusion steps             \\
$\bm{x}_0$, $\bm{x}_n$       & Sample from real data, and the noised data at the $n$-th diffusion step    \\
$\beta_n$          & Noise level at the diffusion step $n$     \\
$\alpha_n$, $\bar{\alpha}_{n}$    & $\alpha_n=1-\beta_n$, $\bar{\alpha}_{n}=\alpha_1 \alpha_2 \ldots \alpha_n$ \\
$f_\phi$                & Volume estimator with parameter $\phi$        \\
$\bm{c}$                & Region features                          \\
$\hat{\bm{s}}$, $\bm{s}$                & Predicted and real flow volume       \\
$\bm{E}_{KG}$                & KG embedding of regions                   \\
$\bm{h}_l$                & Input of region $l$ to the spatial block      \\
$\bm{H}_s$, $\bm{H}_t$                & Output of the spatial temporal block and temporal block     \\
$M_1$                & Number of epochs for pretraining volume estimator     \\
$M_2$  &  We update the volume estimator once every $M_2$ epochs\\
\hline
\end{tabular}
    }
    \vspace{-5px}
    \label{tbl:notations}
\end{table}

\subsection{Denoising Diffusion Probabilistic Model}
\label{sec:DDPM}
Denoising diffusion probabilistic models (DDPM)~\citep{ho2020denoising} are deep generative models that learn a data distribution of variable $x$ by gradually denoising a Gaussian noise. DDPM consists of a forward noising process and a reverse denoising process. 

In the forward process, Gaussian noise is gradually added to the data $\bm{x}_0\sim q(\bm{x}_0)$ to produce a Markov chain:
\begin{equation}
    q(\bm{x}_{1:N}|\bm{x}_0)=\prod_{n=1}^N q(\bm{x}_n|\bm{x}_{n-1}),
\end{equation}
where $q(\bm{x}_n|\bm{x}_{n-1})=\mathcal{N}(\sqrt{1-\beta_t}\bm{x}_{t-1},\beta_t\bm{I})$ and $\beta_n\in(0,1)$ represents the noise level. With the notation $\alpha_n=1-\beta_n$ and $\bar{\alpha}_{n}=\alpha_1 \alpha_2 \ldots \alpha_n$, sampling $\bm{x}_n$ at an arbitrary step $n$ can be written in a close form:
\begin{equation}
    q(\bm{x}_n|\bm{x}_0)=\mathcal{N}(\bm{x}_n;\sqrt{\bar{\alpha}}\bm{x}_0,(1-\bar{\alpha}_n)\bm{I}),
\end{equation}
which can also be reparameterized as 
\begin{equation}
\label{eqn:sample_xn}
    \bm{x}_n=\sqrt{\bar{\alpha}_n}\bm{x}_0+\sqrt{1-\bar{\alpha}_n}\bm{\epsilon}
\end{equation}
with $\bm{\epsilon}$ sampled from a standard gaussian noise $\bm{\epsilon}\sim\mathcal{N}(\bm{0},\bm{I})$.

The reverse process recurrently denoises pure Gaussian noise $\bm{x}_T\sim\mathcal{N}(\bm{0},\bm{I})$ to recover the original data $\bm{x}_0$:
\begin{equation}
\begin{split}
\label{eqn:reverse_process}
    p_\theta(\bm{x}_{0:N})=&p(\bm{x}_n)\prod_{n=1}^N p_\theta(\bm{x}_{n-1}|\bm{x}_n),\\
    p_\theta(\bm{x}_{n-1}|\bm{x}_n)=&\mathcal{N}(\bm{x}_{n-1};\bm{\mu}_\theta(\bm{x}_n,n),\sigma_\theta(\bm{x}_n,n)\bm{I}).
\end{split}   
\end{equation}
To obtain $\bm{x}_{n-1}$ from $\bm{x}_n$, we train a denoising network $\bm{\epsilon}_\theta$ to predict the noise $\bm{\epsilon}$ added to $\bm{x}_0$ using $\bm{x}_n$. Then, $\bm{x}_0$ can be calculated by Equation~\ref{eqn:sample_xn} as $\bm{x}_0=\frac{1}{\sqrt{\bar{\alpha}_n}}(\bm{x}_n-\sqrt{1-\bar{\alpha}_n}\bm{\epsilon}_\theta(\bm{x}_n,n)$). On the other hand, we have the forward process posteriors:
\begin{equation}
\label{eqn:q_posterior}
    q(\bm{x}_{n-1}|\bm{x}_n,\bm{x}_0)=\mathcal{N}(\bm{x}_{n-1};\tilde{\bm{\mu}}(\bm{x}_n,\bm{x}_0),\tilde{\beta}_n\bm{I}),
\end{equation}
where
\begin{equation}
\begin{split}
    \tilde{\bm{\mu}}(\bm{x}_n,\bm{x}_0)=&\frac{\sqrt{\bar{\alpha}_{n-1}}\beta_n}{1-\bar{\alpha}_{n}}\bm{x}_0+\frac{\sqrt{\alpha_n}(1-\bar{\alpha}_{n-1})}{1-\bar{\alpha}_{n}}\bm{x}_n,\\
    \tilde{\beta}_n=&\frac{1-\bar{\alpha}_{n-1}}{1-\bar{\alpha}_{n}}\beta_n.
\end{split}
\end{equation}
Consequently, we can obtain the parameterization of $p_\theta(\bm{x}_{n-1}|\bm{x}_n)$ in Equation~\ref{eqn:reverse_process}:
\begin{equation}
\begin{split}
    \bm{\mu}_\theta(\bm{x}_n,n)=&\frac{1}{\sqrt{\alpha_n}}(\bm{x}_n-\frac{\beta_n}{\sqrt{1-\bar{\alpha}_n}}\bm{\epsilon}_\theta(\bm{x}_n,n)),\\
    \sigma_\theta(\bm{x}_n,n)=&\frac{1-\bar{\alpha}_{n-1}}{1-\bar{\alpha}_{n}}\beta_n.
\end{split}
\end{equation}
The denoising network $\bm{\epsilon}_\theta$ can be trained by optimizing the following L1 loss function:
\begin{equation}
    \mathcal{L}(\theta)=\mathbb{E}_{\bm{x}_0\sim q(\bm{x}_0),\bm{\epsilon}\sim\mathcal{N}(0,\bm{I}),n}||\bm{\epsilon}-\bm{\epsilon}_\theta(\bm{x}_n,n)||.
\end{equation}

\subsection{Urban Knowledge Graph Construction}
\label{sec:kg_construction}
A KG is a graph that consists of an entity set $\mathcal{E}$, a relation set $\mathcal{R}$ and a fact set $\mathcal{F}$, where each fact is represented as a triplet $(h,r,t)\in\mathcal{F}$, denoting a directional edge from head entity $h$ to tail entity $t$ with relation type $r$. KG has been widely used in urban computing because of its ability to model urban environment as well as capture complex relationships between entities~\citep{wang2021spatio,liu2021improving,liu2021knowledge}.

Inspired by previous works~\citep{liu2022developing,liu2023urbankg,zhou2023hierarchical}, we construct an urban KG (UKG) to model the environmental information in the city and interactions among regions. Specifically, we model the urban regions as entities in UKG, and use relations \textit{BorderBy} and \textit{NearBy} to describe the spatial adjacency relationship between regions because urban flows in nearby regions are easily affected by each other. Furthermore, the flow pattern of a region is also correlated with its function~\citep{xu2016context,lin2019deepstn+}, which is mainly reflected by the point of interests (POIs) in the region. As a result, we add POIs and POI categories as entities into UKG, and use relations \textit{LocateAt} and \textit{CateOf} to describe the location and category of POIs. In addition, we consider the geographical influence and competitive relationships between POIs by relation \textit{CoCheckin} and \textit{Competitive}.
We also calculate the functional similarity for each pair of regions, i.e., the cosine similarity of POI category distribution, and use relation \textit{SimilarFunc} to link regions with similar function.
To further enhance the semantics of urban environment, we model business areas (BA) as entities and their relationships with regions and POIs by \textit{ProvideService} and \textit{BelongTo}.
The details of relations in UKG are shown in Table~\ref{tbl:KG_relations}.

\begin{table}[h]
    \centering
    \vspace{-5px}
    \caption{The details of relations in UKG. BA represents business area.}
    \vspace{-5px}
    \resizebox{0.95\linewidth}{!}{
\begin{tabular}{c|c|l}
\hline
\textbf{Relation}       & \textbf{Head \& Tail   Entity Types} & \textbf{Semantic Information}         \\ \hline
\textit{BorderBy}       & (Region, Region)                     & Regions share part of the boundary    \\
\textit{NearBy}         & (Region, Region)                     & Regions lie within a certain distance \\
\textit{CoCheckin}      & (POI, POI)                           & POIs visited by a user consecutively  \\
\textit{Competitive}    & (POI, POI)                           & Nearby POIs with the same brand       \\
\textit{SimilarFunc}    & (Region, Region)                     & Regions with similar POI distribution \\
\textit{LocateAt}       & (POI, Region)                        & POI locates at the region             \\
\textit{CateOf}         & (POI, Category)                      & Category of POI                       \\
\textit{ProvideService} & (BA, Region)                         & BA covers the region                  \\
\textit{BelongTo}       & (POI, BA)                            & POI locates at the BA                 \\ \hline
\end{tabular}
    }
    \vspace{-5px}
    \label{tbl:KG_relations}
\end{table}

To better make use of the environmental knowledge in UKG, we leverage KG embedding model to learn an embedding vector for each region from UKG. Specifically, we choose a state-of-the-art model TuckER~\citep{balavzevic2019tucker}, which uses Tucker decomposition~\citep{tucker1966some} as the scoring function to measure the plausibility of triplets in UKG:
\begin{equation}
    \phi(h,r,t)=\mathcal{W}\times_1 \bm{e}_h\times_2 \bm{e}_r\times_3 \bm{e}_t,
\end{equation}
where $\mathcal{W}\in\mathbb{R}^{d_{KG}\times d_{KG}\times d_{KG}}$ is a learnable core tensor, $\times_n$ is the tensor product along the $n$-th dimension, and $\bm{e}_h, \bm{e}_r, \bm{e}_t\in \mathbb{R}^{d_{KG}}$ are the embeddings of head entity $h$, tail entity $t$ and relation $r$, respectively. The goal of KG embedding model is to make the scoring function high for triplets that exist in the UKG, and thus the knowledge in UKG can be preserved in KG embeddings.

The benefits of UKG are two fold. First, it integrates various types of entities and relationships in the city comprehensively, and thus the learned embedding of a region can provide a description of the urban environment where it locates. Second, various types of relationships between regions in UKG can help capture the spatial dependencies of urban flow between regions, which will be elaborated in the following section.

\begin{definition}[\textbf{Spatio-temporal Flow Generation on KG}]
The UKG organizes regions in a graph structure, and the urban flow $\bm{F}_t$ of different regions is not independent but connected by various types of relations in the UKG. Consequently, the \textit{urban flow generation} problem can be reformulated as generating a dynamic graph signal for a subset of nodes corresponding to the target regions in the UKG, denoted as $\bm{x}=[\bm{F}_1,\ldots,\bm{F}_T]\in \mathbb{R}^{N_l\times T\times d_f}$, given the node features and graph structure $\mathcal{G}=(\mathcal{E},\mathcal{R},\mathcal{F})$.
\end{definition}

%% file: 4.methods.tex
\section{Methods}
\label{sec:methods}
\begin{figure}[ht]
    \centering
    \vspace{-5px}
    \includegraphics[width=\linewidth]{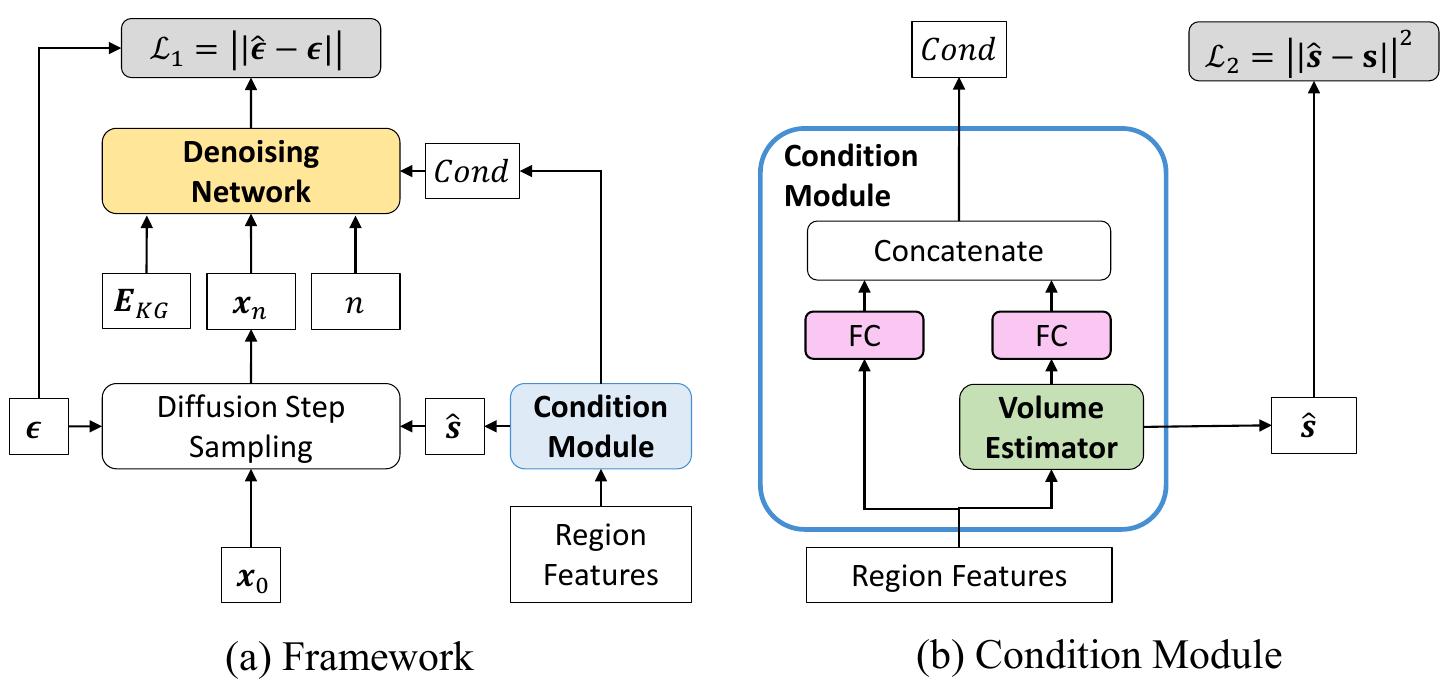}
    \vspace{-5mm}
    \caption{The (a) framework of our model and (b) design of condition module.}
    \label{fig:framework}
    \vspace{-10px}
\end{figure}

Based on the aforementioned DDPM and UKG, we propose a knowledge-enhanced spatio-temporal diffusion model for urban flow generation. The framework of our model is shown in Figure~\ref{fig:framework}(a).
Specifically, we design an auxiliary module to estimate the flow volume of each region, and use the predicted volume to guide the diffusion process for each region separately, which enables learnable and more accurate flow generation process for regions with different volumes.
Moreover, we design a novel knowledge-enhanced denoising network that leverages UKG to capture spatio-temporal dependencies of urban flow and fuse them adaptively based on urban environment.

\subsection{Region Customized Diffusion Process}
\label{sec:customized_diffusion}
The urban flow of a region is closely correlated with its features, such as socioeconomic indicators and demographics~\citep{xu2016context,lin2019deepstn+}. Therefore, we leverage region features to guide the flow generation process through a condition module, as shown in Figure~\ref{fig:framework}(b). Specifically, we design a volume estimator to estimate the flow volume for each region, which is trained together with the diffusion model. On one hand, the estimated flow volume is combined with region features as the condition of the diffusion model. On the other hand, we design a novel learnable and customized diffusion process for each region based on the estimated flow volume.

The volume of urban flow, which we define as the average flow per hour, may vary significantly across different regions. For example, regions in downtown areas usually have a much larger flow volume than suburban regions because of more developed transportation system.
Such spatial variance may lead to the bias of model towards popular regions with larger volume, and thus should be considered directly~\citep{ji2022spatio}.
However, the vanilla diffusion models assume the same endpoint of diffusion process. In other words, the generation process for all regions starts from the same Gaussian noise $\mathcal{N}(0,I)$, which makes it difficult to distinguish regions with different flow volumes. Therefore, we employ the technique proposed in~\citep{han2022card} and make the diffusion process learnable and customized for each region.

Specifically, We design a volume estimator $f_\phi$ with parameters $\phi$ to predict the flow volume based on region features (denoted as $\bm{c}$), and use the predicted volume $f_\phi(\bm{c})$ to guide the diffusion process for each region.
Here $\bm{c}\in\mathbb{R}^{N_l\times d_{fea}}$ represents the features of all target regions, and the output of volume estimator is expanded to the same dimension as $\bm{x}$, i.e., $f_\phi(\bm{c})\in\mathbb{R}^{N_l\times T\times d_f}$.
Then we change the endpoint of diffusion process to Gaussian noise with different mean values:
\begin{equation}
    p(\bm{x}_N|\bm{c})=\mathcal{N}(f_\phi(\bm{c}),\bm{I}).
\end{equation}
Accordingly, with the same notations as in Section~\ref{sec:DDPM}, the forward diffusion process is modified as:
\begin{equation}
\label{eqn:card_forward}
    q(\bm{x}_n|\bm{x}_0,f_\phi(\bm{c}))=\mathcal{N}(\bm{x}_n;\sqrt{\alpha_n}\bm{x}_0+(1-\sqrt{\alpha_n})f_\phi(\bm{c}),(1-\alpha_n)\bm{I}).
\end{equation}
Besides, in the backward denoising process, the posterior in Equation~\ref{eqn:q_posterior} should be changed to:
\begin{equation}
    q(\bm{x}_{n-1}|\bm{x}_n,\bm{x}_0,\bm{c})=\mathcal{N}(\bm{x}_{n-1};\tilde{\bm{\mu}}(\bm{x}_n,\bm{x}_0,f_\phi(\bm{c})),\tilde{\beta}_n\bm{I}),
\end{equation}
where
\begin{equation}
\begin{split}
    \tilde{\bm{\mu}}(\bm{x}_n,\bm{x}_0,f_\phi(\bm{c}))=&
    \frac{\sqrt{\bar{\alpha}_{n-1}}\beta_n}{1-\bar{\alpha}_{n}}\bm{x}_0
    +\frac{\sqrt{\alpha_n}(1-\bar{\alpha}_{n-1})}{1-\bar{\alpha}_{n}}\bm{x}_n\\
    &+(1+\frac{(\sqrt{\bar{\alpha}_{n}}-1)(\sqrt{\alpha_n}+\sqrt{\bar{\alpha}_{n-1}})}{1-\bar{\alpha}_{n}})f_\phi(\bm{c}).
\end{split}
\end{equation}
For the architecture of $f_\phi$, we simply adopt a two-layer feed-forward neural network with Leaky ReLU as the activation function~\citep{han2022card}.
In this way, the start point of flow generation process is approximately the mean value of the flow for each region, which enables more accurate flow generation for regions with different flow volumes.

We consider volume estimation as an auxiliary task and use MSE loss to optimize the volume estimator:
\begin{equation}
    \mathcal{L}_2=||\hat{\bm{s}}-\bm{s}||^2,
\end{equation}
where $\hat{\bm{s}}=f_\phi(\bm{c})$ is the predicted flow volume and $s\in\mathbb{R}^{N_l\times T\times d_f}$ is the real flow volume expanded to the same shape.
In the training process, the volume estimator $f_\phi$ is trained with the denoising network $\bm{\epsilon}_\theta$ in an alternative manner. As shown in Algorithm~\ref{alg:training}, we first pretrain $f_\phi$ for $M_1$ epochs, and then alternatively update the whole diffusion model and $f_\phi$. Specifically, we update the diffusion model for $M_2$ epochs by taking gradient decent step on $\mathcal{L}_1$, and then update $f_\phi$ once with loss function $\mathcal{L}_2$.

\algrenewcommand\algorithmicprocedure{\textbf{Step}}
\algnewcommand{\LineComment}[1]{\State \(\triangleright\) #1}
\begin{algorithm}
\caption{Training of our model}\label{alg:training}
\begin{algorithmic}[1]
\Procedure{1: pretraining $f_\phi$}{}
    \For{$epoch\in\{1,2,\ldots,M_1\}$}
        \State Calculate predicted flow volume $\hat{\bm{s}}=f_\phi(\bm{c})$ 
        \State Update $f_\phi$ using $\mathcal{L}_2=||\hat{\bm{s}}-\bm{s}||^2$
    \EndFor
\EndProcedure
\Procedure{2: joint training $\bm{\epsilon}_\theta$ and $f_\phi$}{}
    \Repeat
        \LineComment{\textbf{Train diffusion model}}
        \For{$epoch\in\{1,2,\ldots,M_2\}$}
            \State Sample flow $\bm{x}_0\sim q(\bm{x}_0)$, noise $\bm{\epsilon}\sim \mathcal{N}(0,I)$
            \State Sample diffusion step $n\sim Uniform(\{1,2,\ldots,N\})$
            \State Calculate $\bm{x}_n$ with Equation~\ref{eqn:card_forward}
            \State Calculate predicted noise $\hat{\bm{\epsilon}}$ with denoising network
            \State Update $\bm{\epsilon}_\theta$ and $f_\phi$ using $\mathcal{L}_1=||\hat{\bm{\epsilon}}-\bm{\epsilon}||$
        \EndFor
        \LineComment{\textbf{Train volume estimator}}
        \State Calculate predicted flow volume $\hat{\bm{s}}=f_\phi(\bm{c})$ 
        \State Update $f_\phi$ using $\mathcal{L}_2=||\hat{\bm{s}}-\bm{s}||^2$
    \Until Convergence
\EndProcedure

\end{algorithmic}
\end{algorithm}

\subsection{Knowledge-enhanced Denoising Network}

\begin{figure*}[h]
    \vspace*{-10px}
    \centering
    \includegraphics[width=.8\linewidth]{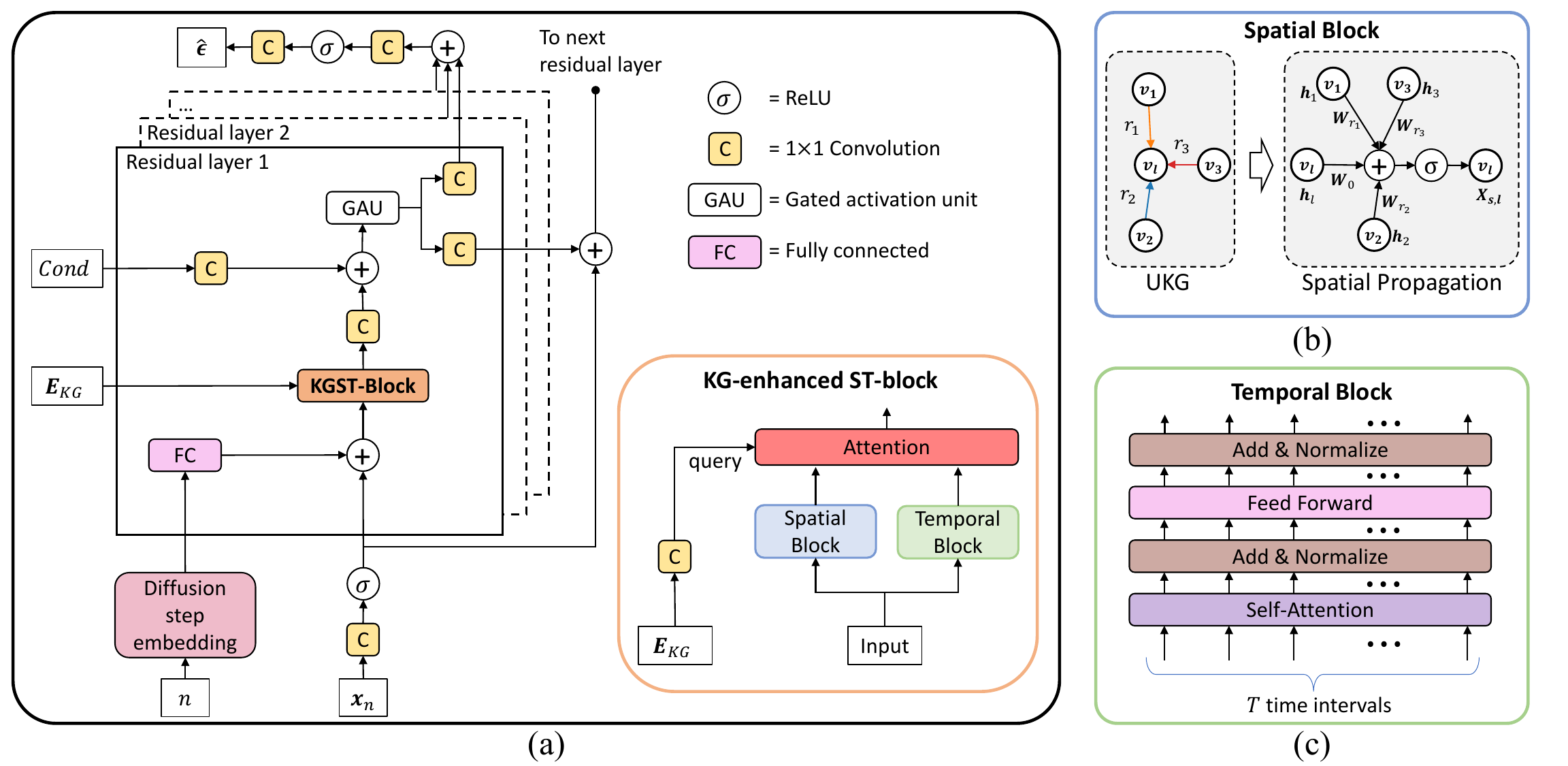}
    \vspace*{-10px}
    \caption{(a) The architecture of denoising network, (b) the detail of spatial block and (c) the detail of temporal block.}
    \label{fig:denoising_network}
    \vspace*{-10px}
\end{figure*}

The core of DDPM lies in the denoising network $\bm{\epsilon}_\theta$, which aims to predict the noise $\bm{\epsilon}$ from noised data $\bm{x}_n$.
The architecture of denoising network is usually designed according to the specific task. Previous works about sequence generation typically use WaveNet-based structure~\citep{oord2016wavenet,kong2020diffwave}, which shows great performance in modeling time series while it cannot capture spatial dependencies in urban flow data. To bridge the gap, we propose a novel knowledge-enhanced network empowered by UKG to capture spatio-temporal dependencies of urban flow for better denoising ability.

\subsubsection{Denoising Network Architecture}
The architecture of our denoising network is shown in Figure~\ref{fig:denoising_network}. Specifically, we adopt the DiffWave architecture as the backbone, which consists of multiple residual layers. It takes noised data $\bm{x}_n\in\mathbb{R}^{N_l\times T\times d_f}$, diffusion step $n$, KG embeddings of regions $\bm{E}_{KG}$ and conditions as input, and output the predicted noise:
\begin{equation}
    \hat{\bm{\epsilon}}=\bm{\epsilon}_\theta(\bm{x}_n,n,\bm{E}_{KG},Cond).
\end{equation}
The output $\hat{\bm{\epsilon}}\in\mathbb{R}^{N_l\times T\times d_f}$ is of the same shape as $\bm{x}_0$ and $\bm{\epsilon}$.

Specifically, we first use 128-dimensional diffusion step embedding for diffusion step $n$ in the diffusion step embedding block:
\begin{equation}
\begin{split}
    n_{embedding}=[sin(10^{0\times4/63}n),\ldots,sin(10^{63\times4/63}n),\\
    cos(10^{0\times4/63}n),\ldots,cos(10^{63\times4/63}n)],
\end{split}
\end{equation}
followed by two fully connected layers before being fed into residual layers.
In the meanwhile, the input noised data $\bm{x}_n$ is mapped by a Conv$1\times1$ and sent to the first residual layer.
In each residual layer, we first combine the input with projected diffusion step embedding to obtain flow representation $\bm{h}\in\mathbb{R}^{N_l\times T\times d_h}$, and then design a KG-enhanced ST-block (KGST-Block) to capture spatio-temporal dependencies of urban flow with the guidance of KG embedding, i.e., $\bm{H}_{st}=KGST(\bm{h},\bm{E}_{KG})$, which will be elaborated in Section~\ref{sec:KGST_Block}. The conditioner of each region $Cond$ is added to the output of KGST-Block $\bm{H}_{st}$. After a gated activation unit, part of the output is connected to the next residual layer as input, while the rest is added to the final output through skip connection.
The Conv$1\times1$ blocks in the network are used to map data to proper dimensions. Finally, the output $\hat{\bm{\epsilon}}$ is the sum of data through skip-connections from each residual layer after projection by two Conv$1\times1$ blocks.
We then introduce the details of KGST-Block.

\subsubsection{KG-enhanced ST-Block Design}
\label{sec:KGST_Block}
The KGST block consists of a spatial block and a temporal block for modeling spatial and temporal dependencies, after which an attention block is used for spatio-temporal fusion with the guidance of KG embeddings.

For the spatial block, we adopt 1-layer R-GCN~\citep{schlichtkrull2018modeling} to capture the spatial dependency of urban flow among regions. R-GCN is a kind of graph convolutional network that aggregates information from the neighborhood of entities through different relations separately. As a result, the flow information of regions that are connected in UKG, i.e., nearby regions and functionally similar regions, can be propagated through the graph. Note that here we only use the subgraph of UKG that consists of training or testing regions and relations among them. Specifically, the output of R-GCN can be obtained as:
\begin{equation}
    \bm{H}_{s,l}=\sigma(\sum_{r\in \mathcal{R}}\sum_{j\in \mathcal{N}_l^r}\bm{W}_r\bm{h}_j+\bm{W}_0\bm{h}_l),
\end{equation}
where $\bm{h}_l\in\mathbb{R}^{T\times d_h}$ is the input of region $l$, $\mathcal{N}_l^r$ is the set of regions connected to region $l$ via relation $r$ and $\bm{W}_r$, $\bm{W}_0\in\mathbb{R}^{d_h\times d_h}$ are learnable weight matrices. $\sigma$ is a nonlinear activation function. 

In the temporal block, we use Transformer layer~\citep{vaswani2017attention} to model the temporal dependency of urban flow, which consists of a multi-head attention layer, a fully-connected layer, and layer normalization.

For different regions, the impact of spatial and temporal flow dependencies may be different depending on the urban environment. For instance, regions in downtown areas usually have a large flow transition with nearby regions, and thus the spatial dependency should be emphasized. However, the flow patterns in suburban residential regions exhibit strong temporal patterns as people usually leave for work in the morning and go home in the evening, while the flow there is not influenced much by nearby regions.

\begin{table*}[t]
    \centering
    \caption{The basic information of four real-world datasets and basic statistics of UKGs.}
    \vspace*{-10px}
    \resizebox{0.8\textwidth}{!}{
\begin{tabular}{cc|cccc}
\hline
\textbf{}                                                                                  & \textbf{Location}        & New York City                                                         & Beijing                                                               & Washington, D.C.                                                      & Baltimore                                                             \\ \hline
\multirow{6}{*}{\textbf{\begin{tabular}[c]{@{}c@{}}Basic\\      information\end{tabular}}} & \textbf{Flow type}       & inflow \& outflow                                                     & inflow \& outflow                                                     & inflow                                                                & inflow                                                                \\
 & \textbf{Time span}       & \begin{tabular}[c]{@{}c@{}}2016.01.01-\\      2016.06.30\end{tabular} & \begin{tabular}[c]{@{}c@{}}2018.01.01-\\      2018.01.31\end{tabular} & \begin{tabular}[c]{@{}c@{}}2019.01.01-\\      2019.05.31\end{tabular} & \begin{tabular}[c]{@{}c@{}}2019.01.01-\\      2019.05.31\end{tabular} \\
 & \textbf{Time interval}   & 1 hour                                                                & 1 hour                                                                & 1 hour                                                                & 1 hour                                                                \\
            & \textbf{Flow length}     & 24 hours                                                              & 24 hours                                                              & 24 hours                                                              & 24 hours                                                              \\
         & \textbf{\#Train regions} & 219                                                                   & 648                                                                   & 178                                                                   & 193                                                                   \\
            & \textbf{\#Test regions}  & 61                                                                    & 362                                                                   & 59                                                                    & 210                                                                   \\ \hline
\multirow{3}{*}{\textbf{\begin{tabular}[c]{@{}c@{}}UKG\\      statistics\end{tabular}}}    & \textbf{\#Entities}      & 26,604                                                                & 36,752                                                                & 11,988                                                                & 16,579                                                                \\
 & \textbf{\#Relations}     & 5                                                                     & 9                                                                     & 5                                                                     & 5                                                                     \\
& \textbf{\#Facts}         & 67,418                                                                & 149,350                                                               & 33,112                                                                & 47,884                                                                \\ \hline
\end{tabular}
    }
    \vspace{-10px}
    \label{tbl:datasets_info}
\end{table*}

Therefore, we utilize the KG embeddings of regions with environmental information preserved to guide the fusion of spatial and temporal dependencies. Specifically, we leverage the attention mechanism with KG embedding as the query. For each region $l$, let $\bm{H}_{s,l}$, $\bm{H}_{t,l}\in\mathbb{R}^{T\times d_h}$ denote the representation after spatial and temporal block respectively, and $\bm{E}_{KG,l}\in\mathbb{R}^{d_{KG}}$ be its KG embedding. We first project the vectors to get the query vector $\bm{Q}=\bm{E}_{KG,l}\bm{W}_Q$, the key vector $\bm{K}_{i}=\bm{H}_{i,l}\bm{W}_K$ and the value vector $\bm{V}_i=\bm{H}_i\bm{W}_V$ ($i\in\{s,t\}$). Then we calculate the importance of spatial and temporal representations as:
\begin{equation}
    \alpha_i=softmax(\frac{\bm{Q}\bm{K}_i^{\top}}{\sqrt{d_h}}), i\in\{s,t\}.
\end{equation}
Finally, the output is calculated as
\begin{equation}
    \bm{H}_{st,l}=(\alpha_s \bm{V}_s+\alpha_t \bm{V}_t)\bm{W}_O,
\end{equation}
where $\bm{W}_O$ is the output projection matrix.

%% file: 5.experiments.tex
\section{Experiments}
\label{sec:experiments}

\subsection{Experiment Settings}
\subsubsection{Datasets}
We conduct experiments on four real-world datasets to examine the effectiveness of our model. We construct a UKG for each dataset according to methods in Section~\ref{sec:kg_construction}, and the basic information of datasets and the statistics of corresponding UKGs are reported in Table~\ref{tbl:datasets_info}. Due to the lack of data in NYC, D.C., and Baltimore datasets, we do not incorporate business areas and several relations, i.e., \textit{CoCheckin}, \textit{Competitive}, \textit{ProvideService} and \textit{BelongTo}, in their UKGs.

Each dataset contains the hourly urban flow of all regions in multiple days, and we use each day (24 hours) as a training sample. Since the flow patterns are quite different on weekdays and weekends, we only use the data on weekdays in our experiments, and our method can be directly applied to the weekends in the same way. We present the details of datasets in Appendix~\ref{app:dataset}.

\subsubsection{Baselines}
\label{sec:baselines}
Since there are no other works that use the same setting as ours, we compare the performance with the following conditional deep generative models. To ensure fair comparison, we use the same condition as our model to generate flow for each region, i.e., region features and KG embeddings.
\begin{itemize}[leftmargin=10px]
    \item \textbf{Conditional GAN}~\citep{mirza2014conditional}: It is the conditional version of vanilla GAN, whose generator and discriminator are both MLPs.
    \item \textbf{DGAN}~\citep{lin2020using}: It is a GAN-based model that first generates metadata and then generated time series conditioned on metadata. In our experiments, we omit the metadata generation process and use our condition as metadata.
    \item \textbf{RCGAN}~\citep{esteban2017real}: It uses conditional GAN structure with LSTM as generator and discriminator to generate time series.
    \item \textbf{CVAE}~\citep{buehler2020data}: It is a kind of conditional variational autoencoder (VAE) used for financial time series generation conditioned on market states. We adapt it for urban flow generation conditioned on region features.
    \item \textbf{Diffwave}~\citep{kong2020diffwave}: It is a diffusion-based model for conditional waveform generation. The denoising network of it is similar to ours while it does not consider spatial dependency among regions.
\end{itemize}

In addition, we also compare with several state-of-the-art urban flow prediction models.
\begin{itemize}[leftmargin=10px]
    \item \textbf{STGCN}~\citep{yu2018spatio}: It employs graph convolution layers and gated convolution layers to capture spatial and temporal correlations, respectively.
    \item \textbf{CCRNN}~\citep{ye2021coupled}: It proposes a novel graph convolution architecture for travel demand prediction where the adjacency matrices in different layers are self-learned during training, and it adopts GRU to capture temporal dynamics.
    \item \textbf{MSDR}~\citep{liu2022msdr}: It uses a new variant of RNN that consider multiple historical time step to capture temporal dependencies, and combine it with graph-based neural networks for spatio-temporal forecasting.
    \item \textbf{3DGCN}~\citep{xia20213dgcn}: It uses an expanded GCN to capture spatio-temporal correlations of urban flow among regions at different time steps.
\end{itemize}
The urban flow prediction models above need historical flow as input to predict the future flow, which however is not available for test regions. Therefore, we sample the historical flow according to the distribution of train regions, and then generate the next steps recurrently~\citep{lin2020using,hui2022knowledge}.

Furthermore, to evaluate the effectiveness of our denoising network design, we examine a variant of our model with a different denoising network.
\begin{itemize}[leftmargin=10px]
    \item \textbf{KSTDiff(UGnet)}: In this model, we change the denoising network in KSTDiff to UGnet proposed in~\citep{wen2023diffstg}. It uses an Unet-like architecture and adopts Temporal Convolution Network (TCN) and graph convolution network (GCN) to capture temporal and spatial dependencies.
    Apart from the denoising network, the framework is the same as ours as shown in Figure\ref{fig:framework}.
\end{itemize}
The implementation details are shown in Appendix~\ref{app:implementation}.

\begin{table*}[htbp!]
    \centering
    \caption{Performance comparison with baselines on four datasets. Best results are presented in bold, and the second best results are underlined.}
    \vspace*{-10px}
    \resizebox{.95\textwidth}{!}{
\begin{tabular}{c|cccc|cccc|cccc|cccc}
\hline
\multicolumn{1}{l|}{} & \multicolumn{4}{c|}{\textbf{NYC}}                                & \multicolumn{4}{c|}{\textbf{Beijing}}                          & \multicolumn{4}{c|}{\textbf{Washington, D.C.}}                   & \multicolumn{4}{c}{\textbf{Baltimore}}                          \\
\textbf{Model}        & \textbf{MAE}   & \textbf{RMSE}  & \textbf{SMAPE} & \textbf{MMD}  & \textbf{MAE}  & \textbf{RMSE} & \textbf{SMAPE} & \textbf{MMD}  & \textbf{MAE}   & \textbf{RMSE}  & \textbf{SMAPE} & \textbf{MMD}  & \textbf{MAE}  & \textbf{RMSE}  & \textbf{SMAPE} & \textbf{MMD}  \\ \hline
STGCN                 & 33.04          & 45.63          & 0.94           & 3.87          & 7.29          & 11.05         & 1.24           & 6.32          & 25.56          & 59.19          & 1.16           & 5.33          & 14.34         & 30.78          & 0.99           & 4.81          \\
CCRNN                 & 37.78          & 46.73          & 0.93           & 6.02          & 7.47          & 11.18         & 1.25           & 6.35          & 31.98          & 57.01          & 1.12           & 6.17          & 13.72         & 28.21          & 0.93           & 3.70          \\
MSDR                  & 32.98          & 43.31          & 0.81           & 3.36          & 6.87          & 11.27         & 1.22           & 5.06          & 32.45          & 58.30          & 1.12           & 5.38          & 15.28         & 29.32          & 1.04           & 4.16          \\
3DGCN                 & 27.82          & 41.38          & 0.80           & 3.94          & 4.37          & 10.58         & 1.08           & 3.51          & 22.44          & 56.68          & 0.91           & {\ul 2.14}    & 13.89         & 29.92          & {\ul 0.92}     & {\ul 2.86}    \\ \hline
Conditional GAN       & 26.75          & 41.73          & 0.71           & 4.88          & 3.93          & 9.98          & 1.04           & 4.46          & 19.51          & 54.43          & 0.86           & 4.31          & 35.90         & 87.03          & 1.02           & 4.80          \\
DGAN                  & 52.90          & 81.95          & 0.81           & 5.24          & 8.14          & 25.93         & 1.23           & 4.34          & 29.12          & 95.16          & 1.31           & 4.55          & 12.92         & 26.94          & 1.38           & 4.09          \\
CVAE                  & 36.51          & 58.48          & 0.70           & 4.19          & 2.67          & 6.61          & {\ul 0.74}     & 4.17          & 28.76          & 64.30          & 1.82           & 4.17          & 82.97         & 254.28         & 0.97           & 4.50          \\
RCGAN                 & {\ul 23.07}    & {\ul 36.12}    & {\ul 0.67}     & 5.09          & 4.11          & 10.47         & 1.21           & 5.45          & {\ul 18.35}    & 48.28          & {\ul 0.82}     & 2.40          & 40.54         & 128.67         & 1.08           & 5.68          \\
Diffwave              & 24.53          & {\ul 36.12}    & 0.68           & {\ul 3.20}    & 2.30          & 6.37          & 0.78           & 1.85          & 36.82          & 51.07          & 1.19           & 5.70          & {\ul 12.88}   & {\ul 21.26}    & 1.05           & 3.71          \\ \hline
KSTDiff(UGnet)            & 34.04          & 51.35          & 0.75           & 3.62          & {\ul 2.26}    & {\ul 5.31}    & {\ul 0.74}     & {\ul 1.82}    & 28.17          & {\ul 46.48}    & 0.97           & 3.91          & 20.73         & 40.01          & 1.11           & 3.26          \\
KSTDiff                  & \textbf{19.46} & \textbf{32.07} & \textbf{0.52}  & \textbf{2.58} & \textbf{1.96} & \textbf{5.15} & \textbf{0.65}  & \textbf{1.43} & \textbf{15.67} & \textbf{41.83} & \textbf{0.65}  & \textbf{1.79} & \textbf{8.35} & \textbf{15.56} & \textbf{0.68}  & \textbf{2.00} \\
Improvement           & 15.6\%         & 11.2\%         & 22.4\%         & 19.4\%        & 13.3\%        & 3.0\%         & 12.2\%         & 21.4\%        & 14.6\%         & 10.0\%         & 20.7\%         & 16.4\%        & 35.2\%        & 26.8\%         & 26.1\%         & 30.1\%        \\ \hline
\end{tabular}
    }
    \label{tbl:results_main}
\end{table*}

\subsubsection{Evaluation Metrics}
We adopt widely used Mean Absolute Error (MAE), Root Mean Square Error (RMSE), and Symmetric Mean Absolute Percentage Error (SMAPE) to measure the performance.
We generate the same number of samples as real data for test regions, and use the mean of all generated samples to calculate MAE, RMSE, and SMAPE.
In addition, we also use maximum mean discrepancy (MMD) to measure the distance between generated data distribution and real data distribution, the details of which are shown in Appendix~\ref{app:metric}.
Specifically, we follow previous work~\citep{esteban2017real} and treat the flows for each region as vectors to calculate MMD, and report the average MMD of all test regions as the final metric.

\subsection{Overall Performance}
The comparison of our model with baselines on four datasets is shown in Table~\ref{tbl:results_main}, from which we have the following observations.

First, our model outperforms all baselines on four datasets by a large margin. For example, our model outperforms existing methods by 14.6\% to 35.2\% in terms of MAE. Such great improvement demonstrates our model's ability to accurately generate urban flow based on region characteristics. 

Second, RCGAN and Diffwave achieve relatively better performance among baselines because they are able to capture temporal dependencies of urban flow. However, they still perform worse than our model mainly because they do not consider spatial dependencies among regions, while our model successfully captures spatio-temporal dependencies through UKG and KGST block. The UGnet architecture is able to model spatio-temporal dependencies by GCN and TCN, while it does not show good performance, which further demonstrates the effectiveness of incorporating urban environmental information by UKG.

Third, the performance of many baselines is sensitive to the data. For example, Diffwave performs almost the second best on NYC, Beijing, and Baltimore datasets, while on D.C. dataset it fails to generate flow accurately. Similarly, the performance of RCGAN is considerable on NYC and D.C. datasets but relatively low on Beijing and Baltimore datasets. The volatile performance may result from that the four real-world datasets have different region features, time spans, and data sources of urban flow. However, our model consistently achieves the best performance on all these datasets, which demonstrates the robustness of our model.

At last, urban flow prediction models generally perform much worse than our model on all datasets. This is because they heavily rely on real historical flow, while the historical flow data on test regions is not available in our setting. The finding demonstrates that predictive modeling of urban flow, though extensively studied, cannot be adapted to data-sparse regions, which further shows the significance of generative modeling of urban flow.

\subsection{Ablation Study}
We conduct an ablation study to further evaluate the influence of region-customized diffusion process~(RCDP) and KGST block in the denoising network. Specifically, to validate the flexibility of adapting the diffusion process to each region, we remove the volume prediction module in Figure~\ref{fig:framework} as well as its training process, and only keep region features in the condition module. In other words, the diffusion process is the same for all regions with $f_\phi(\bm{c})=0$. As for KGST, we remove the spatial block as well as the attention module in the denoising network, and only preserve the temporal block in KGST block. 

\begin{figure}[htbp!]
\vspace*{-10px}
\centering
\hspace{-3mm}
    \subfigure[SMAPE]{
    {\label{subfig:abl_KGvol_smape}}
    \includegraphics[width=.48\linewidth]{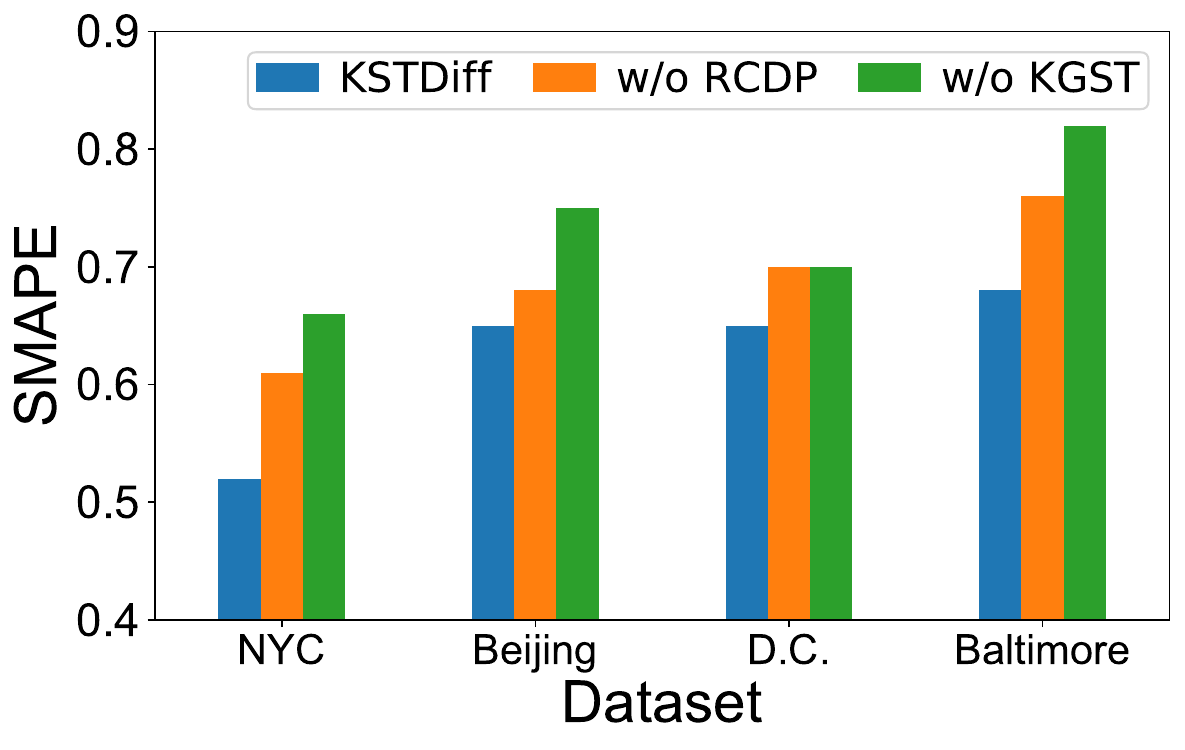}
    }
    \subfigure[MMD]{
    {\label{subfig:abl_KGvol_mmd}}
    \includegraphics[width=.48\linewidth]{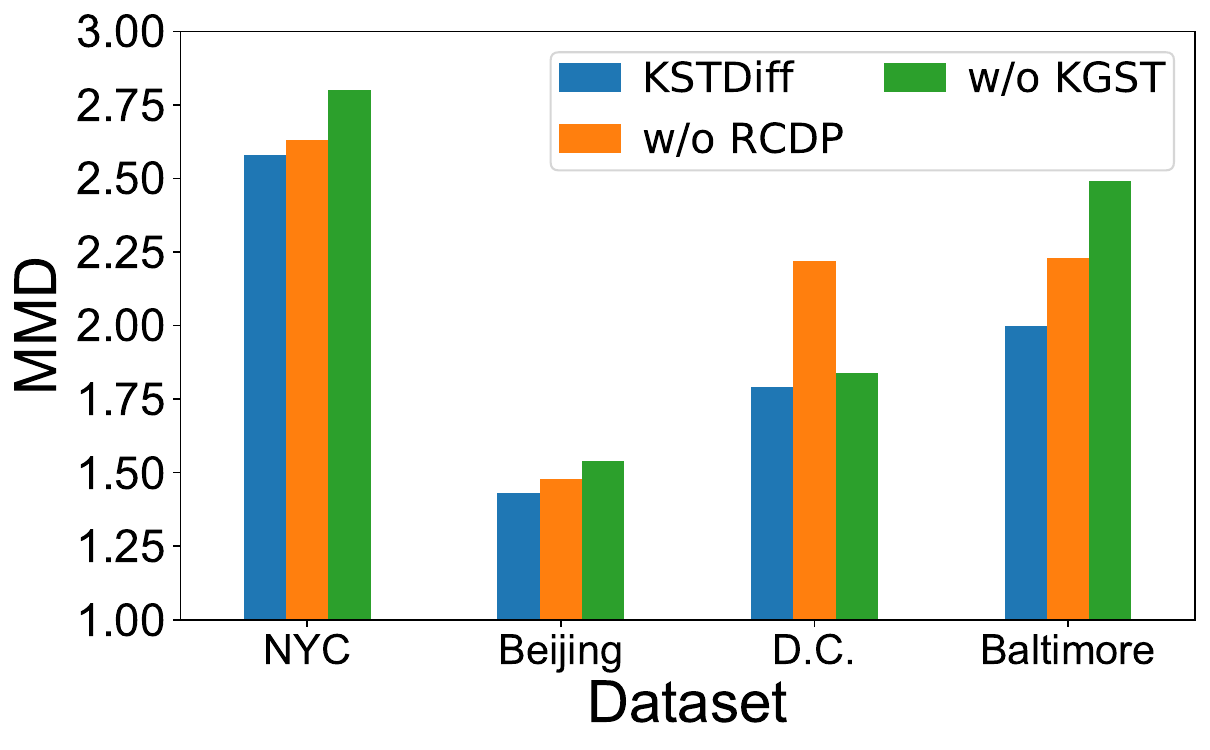}
    }
\hspace{-3mm}
\vspace*{-5px}
\caption{Performance comparison of models without KGST block or region-customized diffusion process (RCDP).
}\label{fig:abl_KGvol}
\vspace*{-10px}
\end{figure}

We visualize the SMAPE and MMD metrics on four real-world datasets in Figure~\ref{fig:abl_KGvol}. It can be observed that the performance drops in all cases after omitting RCDP or KGST, which demonstrates the effectiveness of such design. Besides, we find that the performance deteriorates significantly after removing KGST, which suggests that modeling spatial dependencies and urban environment plays an important role in urban flow generation.

\subsection{Case Study}
\subsubsection{Long-term Flow Generation}
\begin{figure}[htbp!]
\vspace*{-10px}
\centering
\hspace{-3mm}
    \subfigure[NYC Dataset]{
    {\label{subfig:long_term_ny_mae}}
    \includegraphics[width=.45\linewidth]{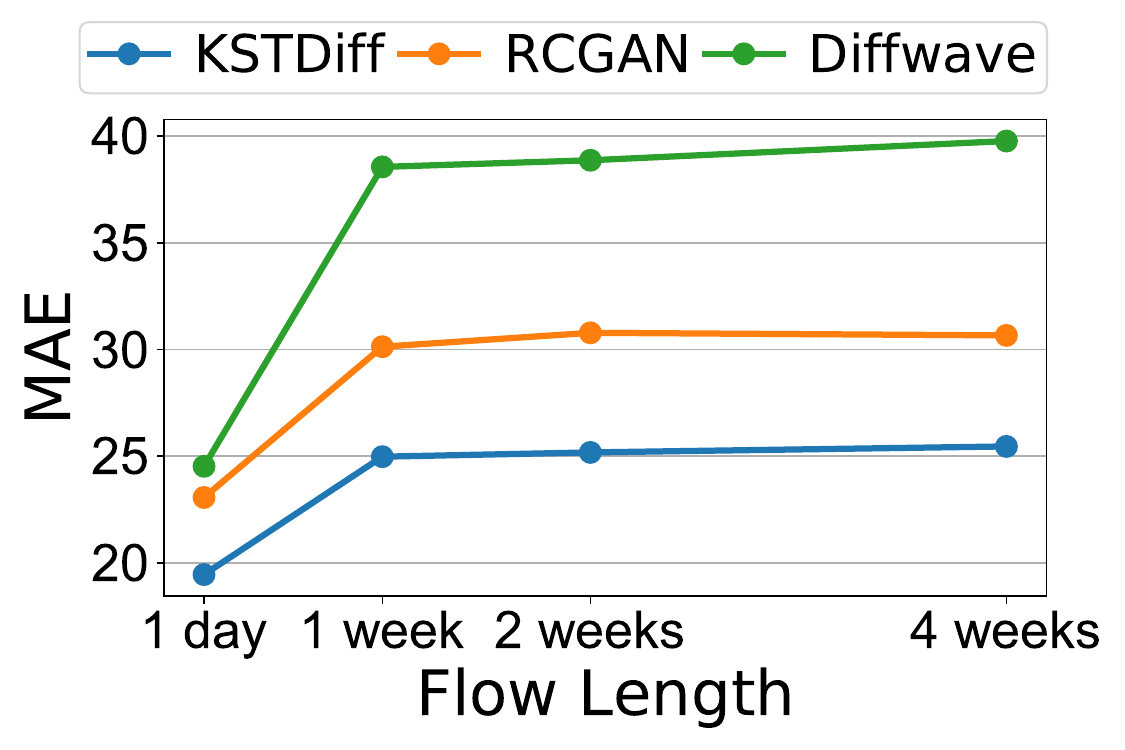}
    }
\hspace{-3mm}
    \subfigure[D.C. Dataset]{
    {\label{subfig:long_term_dc_mae}}
    \includegraphics[width=.45\linewidth]{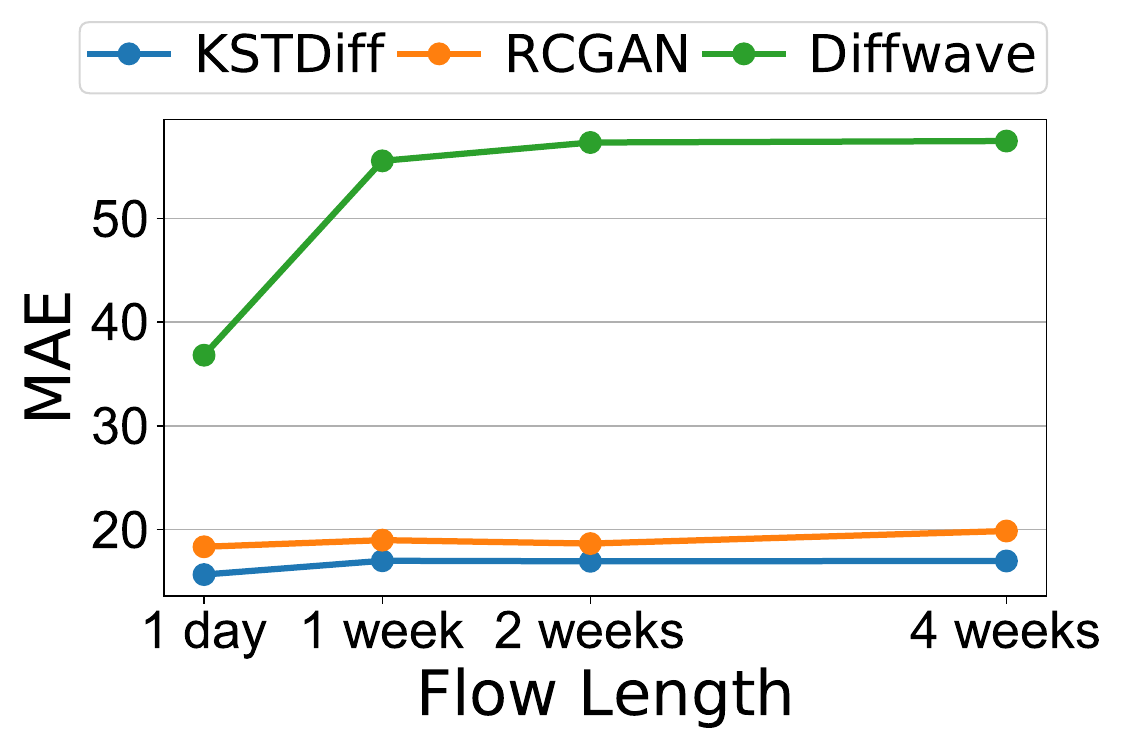}
    }
\hspace{-3mm}
\vspace*{-10px}
\caption{Performance comparison of models for long-term flow generation.}
\label{fig:long_term}
\vspace*{-10px}
\end{figure}

In real-world scenarios, knowing long-term urban flow distribution in advance can better help urban planning~\citep{xie2022multisize}. To evaluate the performance of our model in generating longer flow sequences, we expand the generated flow length to one week, two weeks, and four weeks. For example, we generate the flow with $7\times 24$ time steps for the one-week case. 
We conduct experiments on NYC dataset and D.C. dataset, and compare the performance of our model with the two best baselines, i.e. RCGAN and Diffwave. As shown in Figure~\ref{fig:long_term}, the error generally increases as the flow gets longer, while our model consistently achieves the best performance under various flow lengths, which demonstrates its robustness in modeling urban flow with various lengths. We further present the MMD metric in Appendix~\ref{app:long_term_mmd}.

\subsubsection{Downstream Application}
To evaluate the utility of generated urban flow in supporting other urban flow related tasks, we conduct experiments by using our generated urban flow for OD generation. The OD generation task aims to generate origin-destination (OD) flow among regions, i.e., the number of people moving from one region to another, based on region features and total outflow of each region, which is important for a wide range of applications like transportation planning and epidemics spread modeling~\citep{ren2017efficient,wei2020examining,antoniou2016towards}.

Specifically, we leverage the state-of-the-art OD generation model Deep Gravity~\citep{simini2021deep}, which uses feed-forward neural network with multiple layers to calculate the OD flow between each pair of regions based on their features and outflows. Note that OD generation task does not consider the temporal dynamics of flow, and only aims to generate the total OD flow among regions. Therefore, we use the sum of outflow in all time intervals as the input of Deep Gravity model.
For evaluation, we adopt the same metrics as the original setting including Common Part of Commuters (CPC), Normalized Root Mean Squared Error (NRMSE), Pearson correlation coefficient, and Jenson-Shannon divergence (JSD). Here the benchmark is the performance of Deep Gravity model using real flow data, and a performance closer to the benchmark indicates the better utility of our generated flow.

We compare the performance of OD generation using real outflow data, data generated by our model or baseline models including RCGAN and Diffwave. It can be observed from Table~\ref{tbl:results_deepgravity} that real data achieves the best performance, while the data generated by our model perform closer to real data than other baselines.
Furthermore, the performance of our generated data is not much worse than real data. For example, the CPC of our model is only 7.50$\%$ and 4.81$\%$ lower than the performance using real data on NYC and Beijing datasets, respectively, which demonstrates that our model can be used to help OD generation task in case of lacking real flow data without much performance loss.

\begin{table}[htbp!]
    \centering
    \vspace*{-5px}
    \caption{Performance comparison with baselines in OD generation. Corr. represents Pearson correlation coefficient.}
    \vspace*{-10px}
    \resizebox{.95\linewidth}{!}{
\begin{tabular}{c|cccc|cccc}
\hline
\multicolumn{1}{l|}{} & \multicolumn{4}{c|}{\textbf{NYC}}                                  & \multicolumn{4}{c}{\textbf{Beijing}}                               \\
\textbf{Model}        & \textbf{CPC$\uparrow$}   & \textbf{NRMSE$\downarrow$}  & \textbf{Corr.$\uparrow$} & \textbf{JSD$\downarrow$}   & \textbf{CPC$\uparrow$}   & \textbf{NRMSE$\downarrow$}  & \textbf{Corr.$\uparrow$} & \textbf{JSD$\downarrow$}   \\ \hline
RCGAN                 & 0.656          & 7.87\%          & 0.634          & 0.283          & 0.425          & 0.96\%          & 0.503          & 0.468          \\
Diffwave              & 0.671          & 8.19\%          & 0.604          & 0.296          & 0.540          & 0.80\%          & 0.638          & 0.410          \\
KSTDiff                  & {\ul 0.715}    & {\ul 7.39\%}    & {\ul 0.683}    & {\ul 0.258}    & {\ul 0.554}    & {\ul 0.79\%}    & {\ul 0.687}    & {\ul 0.400}    \\
real data             & \textbf{0.773} & \textbf{5.96\%} & \textbf{0.795} & \textbf{0.210} & \textbf{0.582} & \textbf{0.77\%} & \textbf{0.713} & \textbf{0.377} \\ \hline
\end{tabular}
    }
    \vspace*{-10px}
    \label{tbl:results_deepgravity}
\end{table}

\subsubsection{Predictive Modeling of Urban Flow}
The generative modeling of urban flow essentially aims to learn its conditional distribution. Therefore, it is possible to adapt our model for predictive modeling of urban flow by using historical flow as the condition.
To validate this hypothesis, we slightly modify our model for urban flow prediction task.
Specifically, in this task, we aim to predict the urban flow for all regions in the next $T_2$ time intervals in the future ($\bm{F}_{t+1},\ldots,\bm{F}_{t+T_2}$) based on urban flow in the past $T_1$ time intervals ($\bm{F}_{t-T_1+1},\ldots,\bm{F}_{t}$), where $T_1$ and $T_2$ are both 12 in our setting. Since historical flow is available in the flow prediction task, we first remove the volume estimator and adopt the vanilla diffusion process without guidance from the predicted volume. Following previous work~\citep{wen2023diffstg}, we consider the combination of historical and future flow $[\bm{F}_{t-T_1+1:t};\bm{F}_{t+1:t+T_2}]$ as the data $x_0$. Moreover, the data $x_0$ with future flow masked $[\bm{F}_{t-T_1+1:t};\bm{0}]$ is served as condition, with which we replace the condition module. The training process is similar to Algorithm~\ref{alg:training} while we no longer need the pretraining and updating of $f_\phi$.

\begin{figure}[htbp!]
\centering
\vspace*{-10px}
\hspace{-3mm}
    \subfigure[D.C. dataset]{
    {\label{subfig:long_term_mmd}}
    \includegraphics[width=.4\linewidth]{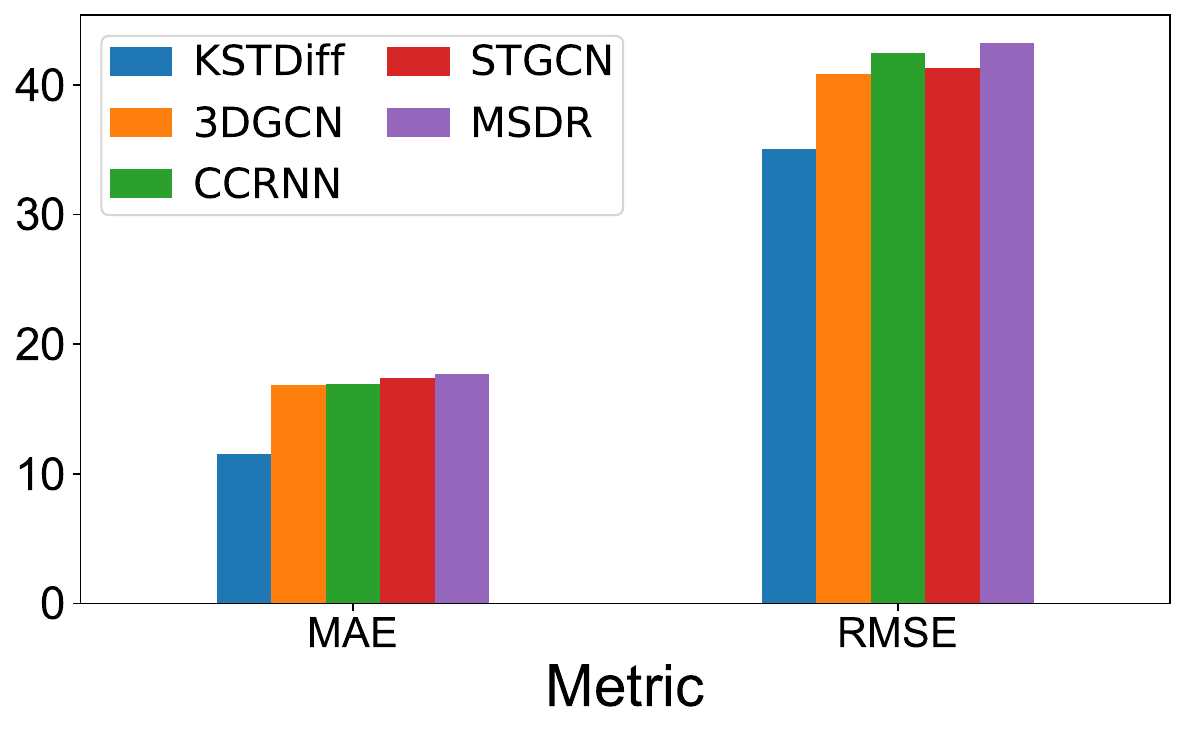}
    }
\hspace{-3mm}
    \subfigure[NYC dataset]{
    {\label{subfig:long_term_mae}}
    \includegraphics[width=.4\linewidth]{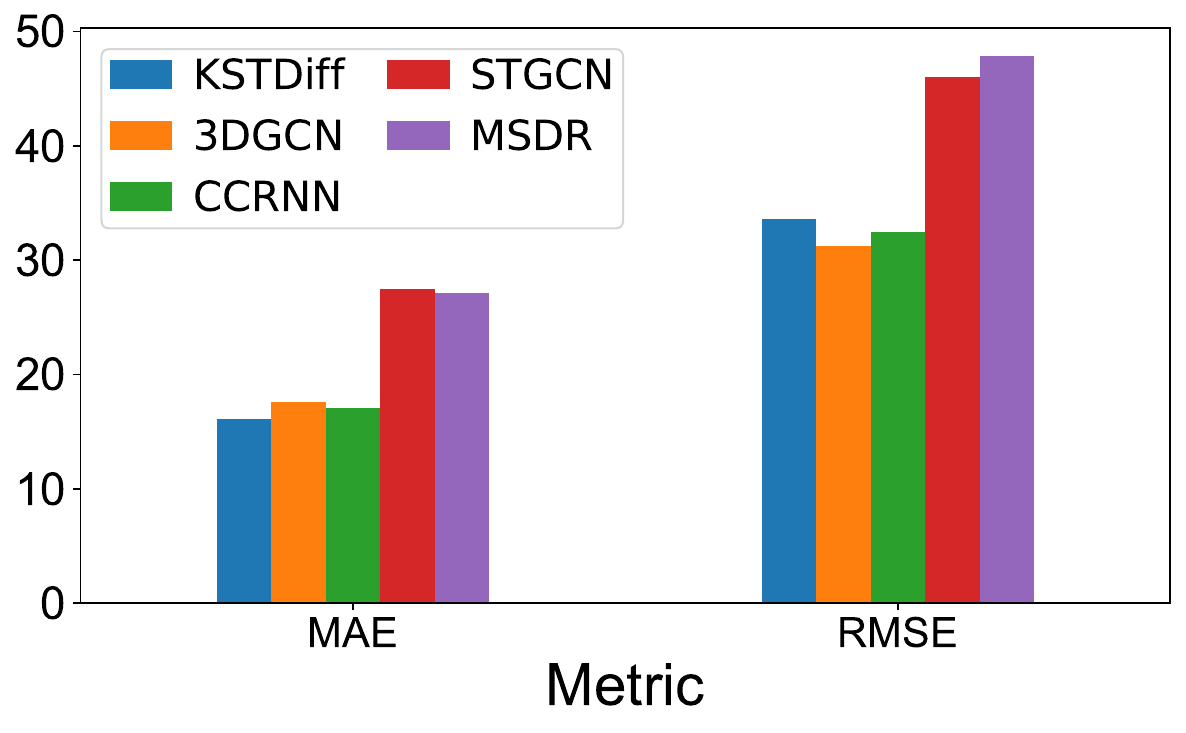}
    }
\hspace{-3mm}
\vspace*{-10px}
\caption{The results of our model and baselines on urban flow prediction task.}
\label{fig:prediction}
\vspace*{-10px}
\end{figure}

We compare the performance of our model with the urban flow prediction baselines mentioned in Section~\ref{sec:baselines}, i.e., 3DGCN, CCRNN, STGCN, and MSDR, on NYC and D.C. datasets.
We use four weeks' data as the training set, one week for validation, and one week for test in chronological order. We choose widely used MAE and RMSE to evaluate the performance and present the results in Figure~\ref{fig:prediction}.
It can be observed that our model outperforms the baselines by 14.2\% in terms of RMSE on D.C. dataset, and achieves comparable performance with the best baselines 3DGCN and CCRNN on NYC dataset (5.5\% improvement in MAE and -7.6\% in RMSE). The findings suggest that our model can be effectively adapted for urban flow prediction, which further demonstrates the power and versatility of generative modeling of urban flow.

%% file: 6.conclusion.tex
\section{Conclusion}
\label{sec:conclusion}

In this paper, we investigate the generative modeling of a typical kind of geospatial data, i.e., urban flow, by studying the urban flow generation problem, which aims to generate urban flow in multiple time steps for regions without historical flow data. We leverage diffusion model to generate urban flow for different regions under different conditions. Specifically, we construct a UKG to model spatial relationships between regions as well as the urban environment, and further propose a knowledge-enhanced spatio-temporal diffusion model to generate urban flow based on UKG. Our model employs a region customized diffusion process to tackle the variance in flow volumes of different regions, and uses a knowledge-enhanced denoising network to capture the spatio-temporal
dependencies of urban flow and the impact of environment. Extensive experiments show the superiority of our model design and the utility of generated flow data. Further studies demonstrate its ability for long-term flow generation and urban flow prediction.
In the future, we aim to improve the efficiency of our model by accelerating the sampling process of diffusion model, e.g., adopting techniques like DPM-Solver~\citep{lu2022dpm}. In addition, a promising direction is to adapt our model to the cross-city scenario that generates urban flow for new cities~\citep{simini2021deep}.

%% file: 7.Appendix.tex
\section{Details of Datasets}
\label{app:dataset}
\begin{figure}[htbp!]
\centering
\hspace{-2mm}
    \subfigure[NYC Dataset]{
    {\label{subfig:ny_traintest}}
    \includegraphics[width=.22\linewidth]{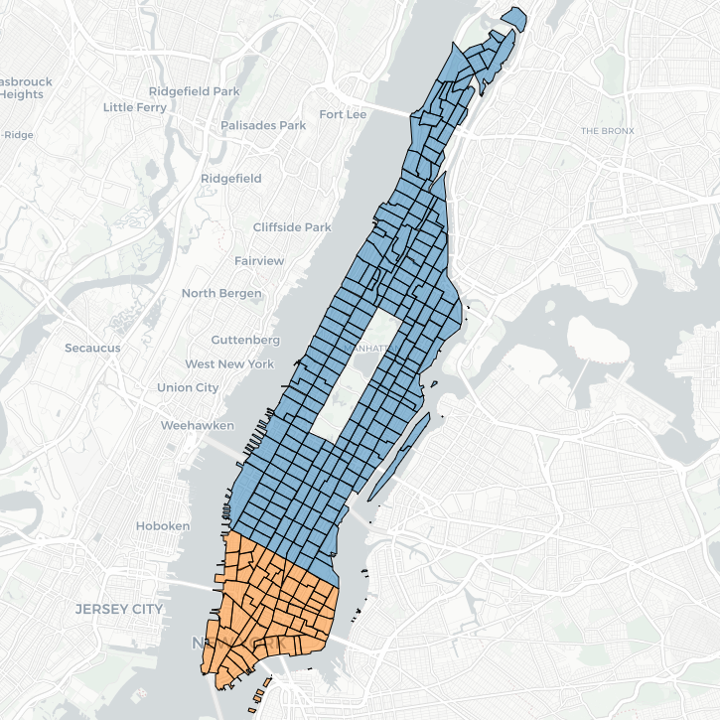}
    }
\hspace{-3mm}
    \subfigure[Beijing Dataset]{
    {\label{subfig:bj_traintest}}
    \includegraphics[width=.22\linewidth]{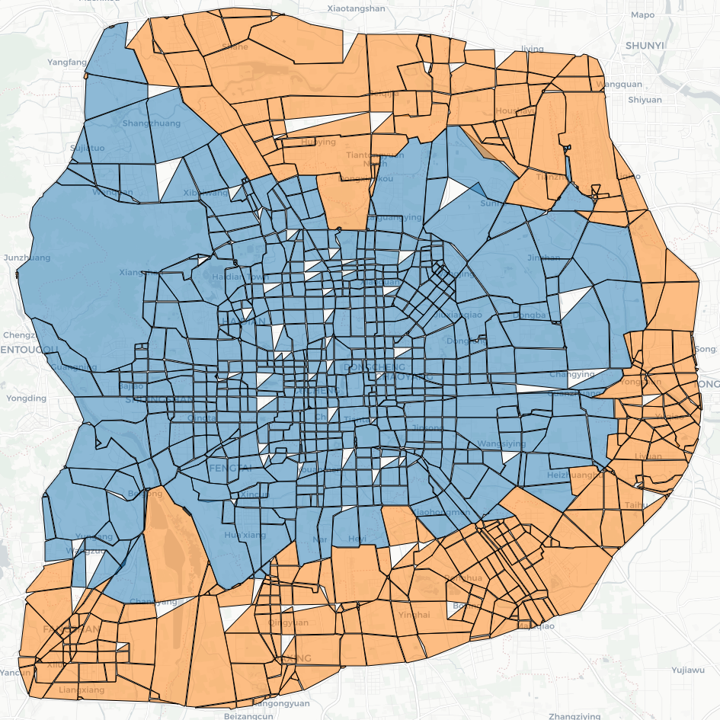}
    }
\hspace{-3mm}
    \subfigure[D.C. Dataset]{
    {\label{subfig:dc_traintest}}
    \includegraphics[width=.22\linewidth]{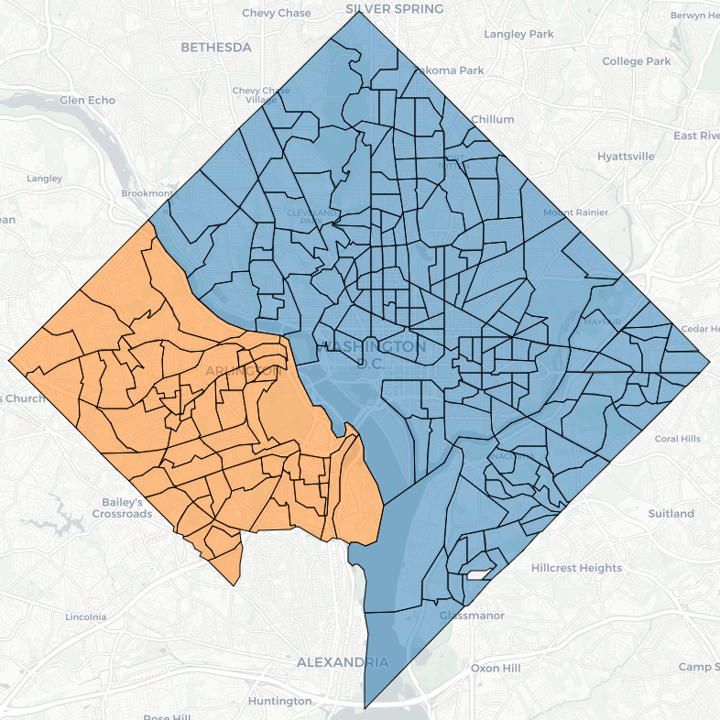}
    }
\hspace{-3mm}
    \subfigure[Baltimore Dataset]{
    {\label{subfig:bm_traintest}}
    \includegraphics[width=.22\linewidth]{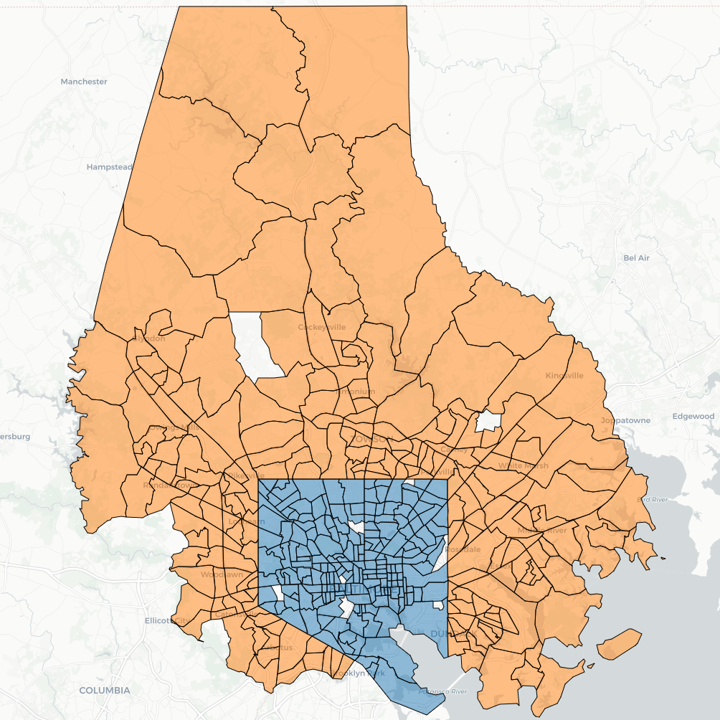}
    }
\caption{The division of train regions and test regions on four datasets. Blue regions denote train regions and orange regions denote test regions.
}\label{fig:traintest}
\end{figure}

In this section, we present the details of four real-world datasets.
\begin{itemize}[leftmargin=10px]
    \item \textbf{NYC Dataset}. In this dataset, we define regions as census tracts and aggregate taxi trips from NYC Open Data\footnote{\url{https://data.cityofnewyork.us/Transportation/2016-Yellow-Taxi-Trip-Data/k67s-dv2t}} to obtain the hourly inflow and outflow. The train and test regions are split based on community districts. Specifically, the Manhattan borough contains 12 community districts, and we choose 9 of them as the train regions and the rest 3 as test regions. The region features include number of POIs, area, population, number of takeaway orders, total price of takeaway orders, number of restaurants, and number of firms.
    \item \textbf{Beijing Dataset}. The regions in Beijing dataset are divided by main road networks. We obtain inflow and outflow for each region within Sixth Ring Road by aggregating user trajectories. As for the selection of train and test regions, we choose regions in downtown areas, i.e., Dongcheng, Xicheng, Chaoyang, Fengtai, Shijingshan, and Haidian districts, as the train regions, and the rest are test regions. The region features consist of number of POIs, population, education level, income level, unemployment rate, and number of crimes.
    \item \textbf{Washington, D.C. Dataset}\footnote{\label{footnote:dc}\url{https://github.com/SonghuaHu-UMD/MultiSTGraph}}. The regions in this dataset are defined as census tracts, and the inflows are calculated by POI-level hourly visits. The train and test regions are split by counties. Specifically, we choose regions in the District of Columbia as train regions, and regions in Arlington County as test regions. This dataset contains 23 region features including demographics, socioeconomic indicators and land use.
    \item \textbf{Baltimore Dataset}\textsuperscript{\ref{footnote:dc}}. The regions and inflows are obtained in the same way as D.C. dataset, and we choose regions in Baltimore City and Baltimore County as train regions and test regions, respectively. This dataset contains the same features as D.C. dataset.
\end{itemize}
The train regions and test regions of four datasets are visualized in Figure~\ref{fig:traintest}. Each region feature is normalized across all regions by z-score normalization.

\section{Implementation}
\label{app:implementation}
In our experiments, we set the number of diffusion steps $N=1000$, use 5 residual layers in the denoising module and the dimension of KG embedding is $d_{KG}=32$. The learning rate is searched from \{1e-4, 5e-4, 1e-3, 5e-3\} and the number of training epochs range from \{200,500,1000,1500,2000\}. The pretraining epoch number $M_1$ is set to 100 and $M_2$ is searched from \{1,10,100,w/o\}, where w/o means that we only pretrain the volume estimator and do not train it anymore in the following training process.
For all models, we tuned the hyperparameters on each of the datasets and report the best results.

\section{Details of Evaluation Metrics}
\label{app:metric}
The maximum mean discrepancy (MMD) is a metric that measures the similarity between two distributions by comparing statistics of the samples. Specifically, for each region, given the generated flow samples $\{f_g^i\}_{i=1}^{n}$ and real flow samples $\{f_r^i\}_{i=1}^{m}$, the MMD is calculated as:
\begin{equation}
\begin{split}
    MMD=&\frac{1}{n(n-1)}\sum_{i=1}^{n}\sum_{j\neq i}^{n}K(f_g^i,f_g^j)
    -\frac{2}{mn}\sum_{i=1}^{n}\sum_{j=1}^{m}K(f_g^i,f_r^j)\\
    &+\frac{1}{m(m-1)}\sum_{i=1}^{m}\sum_{j\neq i}^{m}K(f_r^i,f_r^j),
\end{split}
\end{equation}
where $K(x,y)=exp(-||x-y||^2)/(2\sigma^2)$ is the RBF kernel. 
In our study, we calculate MMD based on the code in this repository\footnote{\url{https://github.com/easezyc/deep-transfer-learning/blob/master/MUDA/MFSAN/MFSAN\_3src/mmd.py}}.

\section{Long-term Flow Generation Results}
\label{app:long_term_mmd}
\begin{figure}[htbp!]
\vspace*{-10px}
\centering
\hspace{-3mm}
    \subfigure[NYC Dataset]{
    {\label{subfig:long_term_ny_mmd}}
    \includegraphics[width=.45\linewidth]{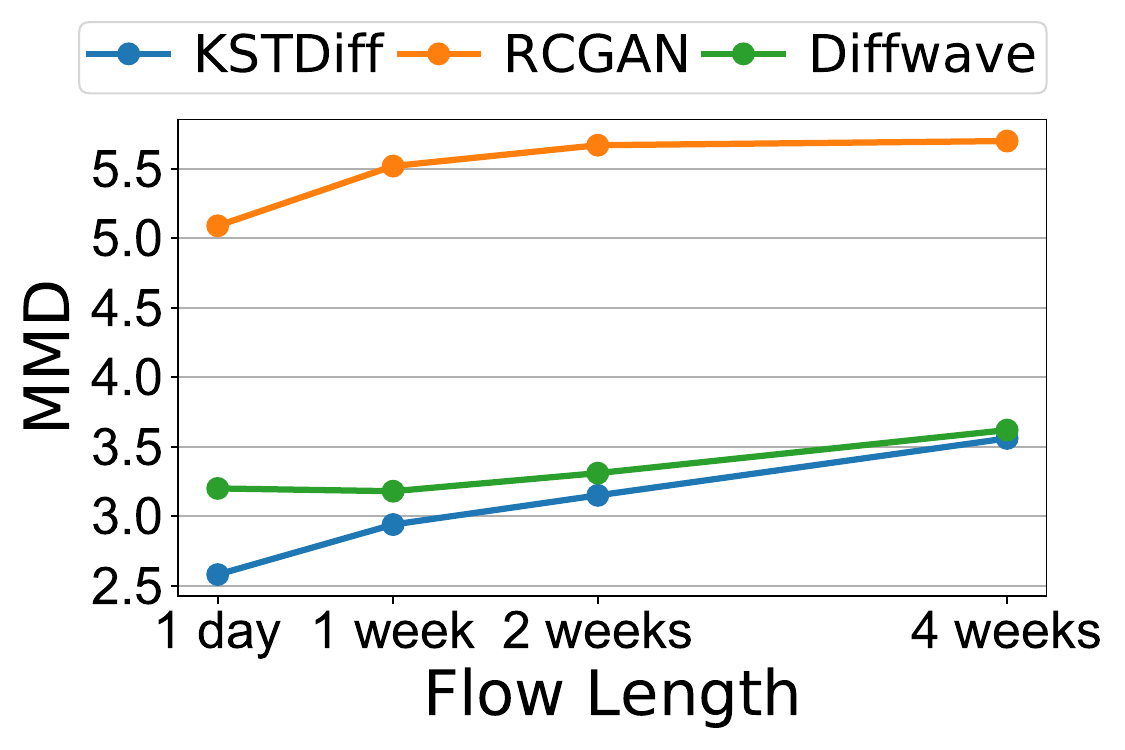}
    }
\hspace{-3mm}
    \subfigure[D.C. Dataset]{
    {\label{subfig:long_term_dc_mmd}}
    \includegraphics[width=.45\linewidth]{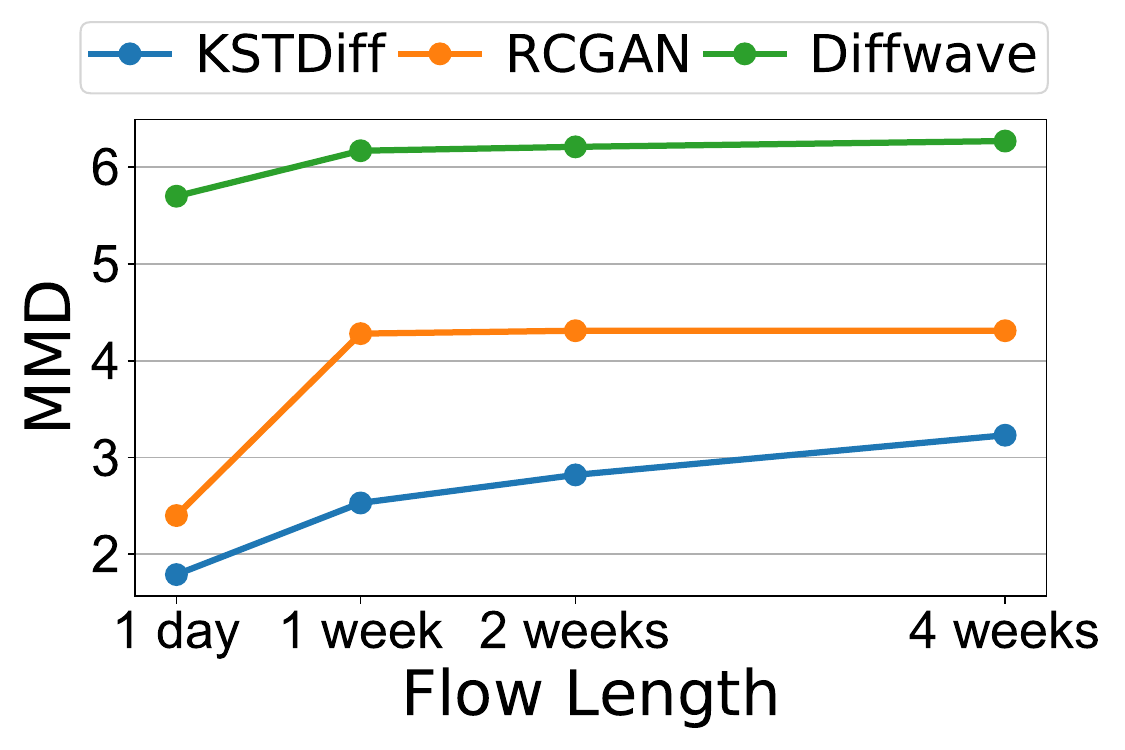}
    }
\hspace{-3mm}
\vspace*{-10px}
\caption{Performance comparison of models for long-term flow generation.}
\label{fig:long_term_mmd}
\vspace*{-10px}
\end{figure}

We present the MMD of long-term flow generation results in Figure~\ref{fig:long_term_mmd}, which shows a similar trend with MAE.
